\documentclass[lettersize,journal]{IEEEtran}
\usepackage{amsmath,amsfonts}
\usepackage{algorithmic}
\usepackage{array}
\usepackage{textcomp}
\usepackage{stfloats}
\usepackage{url}
\usepackage{verbatim}
\usepackage{graphicx}
\def\BibTeX{{\rm B\kern-.05em{\sc i\kern-.025em b}\kern-.08em
    T\kern-.1667em\lower.7ex\hbox{E}\kern-.125emX}}
\usepackage{balance}

\usepackage{algorithmic}
\usepackage{graphicx}
\usepackage{textcomp}
\usepackage{xcolor}
\usepackage{booktabs}
\usepackage{multirow} %
\usepackage{tabularx} 
\usepackage{color}
\usepackage{colortbl}
\usepackage{amsmath} 
\usepackage{algorithm}
\usepackage{algorithmic}
\usepackage{enumitem}
\usepackage{rotating}
\usepackage{subfig}

\usepackage{tikz,xcolor}
\usepackage{bbding}
\usepackage{color}
\usepackage{threeparttable}
\usepackage[implicit=false]{hyperref}
\hypersetup{hidelinks,
	colorlinks=true,
	allcolors=black,
	pdfstartview=Fit,
	breaklinks=true}
\definecolor{lime}{HTML}{A6CE39}
\DeclareRobustCommand{\orcidicon}{
\begin{tikzpicture}
\draw[lime, fill=lime] (0,0)
circle[radius=0.16]
node[white]{{\fontfamily{qag}\selectfont \tiny \.{I}D}};
\end{tikzpicture}
\hspace{-2mm}
}
\foreach \x in {A, ..., Z}{%
\expandafter\xdef\csname orcid\x\endcsname{\noexpand\href{https://orcid.org/\csname orcidauthor\x\endcsname}{\noexpand\orcidicon}}
}

\begin{document}
\title{From Coarse to Fine: Managing Temporal Granularity in Spatio-Temporal Data for Fine-Grained Traffic Prediction}
\author{Shuhao Li\orcidA{}, ~\IEEEmembership{Student Member, ~IEEE}, Weidong Yang\orcidC{}, Yue Cui\orcidB{}, Zizhuo Xu, Lipeng Ma,\\ Fan Zhang, and Xiaofang Zhou\orcidC{},~\IEEEmembership{Fellow,~IEEE}

\thanks{Manuscript received November 25, 2025.}
\thanks{Shuhao Li, Weidong Yang, and Lipeng Ma with the College of Computer Science and Artificial Intelligence, Fudan University, Shanghai 200082, China (shli23@m.fudan.edu.cn)

Yue Cui is with Tongyi Lab, Alibaba Group, Cloud Valley, NO.1008 Dengcai Street, Xihu District, Hangzhou 310020, China

Zizhuo Xu and Xiaofang Zhou are with The Hong Kong University of Science and Technology, Hong Kong SAR 999077, China 

Fan Zhang is with Guangzhou University, Guangzhou 510006, China 
}
}

\markboth{Journal of \LaTeX\ Class Files,~Vol.~18, No.~9, September~2020}%
{How to Use the IEEEtran \LaTeX \ Templates}

\maketitle

\begin{abstract}
Efficient acquisition, storage, and utilization of traffic data are critical challenges in spatio-temporal data management. Most traffic data systems collect and store observations at fixed, coarse-grained temporal intervals to reduce storage and computation costs. However, such coarse-grained data severely limits downstream applications that require predictions at a finer temporal granularity. Collecting and maintaining fine-grained traffic data across all locations and time periods would impose a substantial burden on database storage and preprocessing pipelines. To address this temporal granularity mismatch, we formulate a novel problem: predicting fine-grained future traffic using coarse-grained sampled data. We propose the \underline{\textbf{S}}patial-\underline{\textbf{T}}emporal \underline{\textbf{R}}efinement \underline{\textbf{P}}redictor (\textbf{STRP}), a granularity-aware framework for spatio-temporal data systems. STRP integrates two components: Tree Convolution for efficient and interpretable spatial dependency modeling, and Inverse Dilated Convolution for progressive temporal extrapolation. STRP supports two practical prediction settings—window-based and duration-based—to handle different forms of granularity mismatch. Experiments on six benchmark datasets show that STRP significantly outperforms state-of-the-art baselines in both accuracy and efficiency. Our work offers a practical and interpretable approach to managing granularity mismatches in spatio-temporal traffic data systems. 
\end{abstract}


\section{Introduction}
\begin{figure}[t]
\centerline{\includegraphics[width=0.45\textwidth]{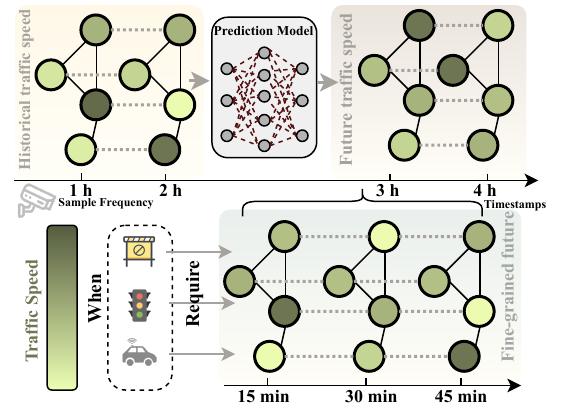}}
\caption{Example of fine-grained future prediction illustrates that trned on 1-hour sampling data cannot predict 15-minute intervals.}
\vspace{-0.5cm}
\label{fig: intro}
\end{figure}

\IEEEPARstart{I}n modern spatio-temporal data systems, urban traffic is continuously monitored via fixed sensors and vehicle trajectories to support routing, congestion analysis, and intelligent transportation systems \cite{wang2021prediction, shao2020eco}. To balance scalability and storage, mainstream deployments adopt fixed, coarse-grained temporal sampling and treat each slice as the basic unit for data management and predictive modeling \cite{cao2024spatiotemporal, guo2021hierarchical}. Consequently, much of the literature emphasizes architectural refinements to improve prediction accuracy and representation capacity \cite{zhao2023causal, ouyang2024trafficgpt, jiang2023spatio, li2024st}. While effective for long-term trends, this practice limits high-frequency, time-sensitive forecasting in operational settings. As illustrated in \textbf{Figure~\ref{fig: intro}}, models trained on hourly histories lack the temporal resolution to produce fine-grained 5- or 15-minute forecasts, which undermines downstream applications that require real-time responsiveness, such as signal control and autonomous lane-changing \cite{wang2024traffic, li2024deep}.

Although collecting and storing fine-grained data across all temporal and spatial dimensions is technically possible, it incurs substantial overhead on storage, preprocessing, and indexing, and maintenance, rendering such a design impractical at scale. To resolve this temporal granularity mismatch, we define the fine-grained traffic prediction problem: using only coarse-grained histories, intelligently recover and forecast fine-grained future states without inflating system cost. This yields a deployable solution for time-critical applications and meaningfully expands the predictive capacity of existing large spatio-temporal data systems.

While the proposed problem broadens the scope of spatio-temporal prediction, it also introduces a set of challenges that span both algorithmic complexity and system-level constraints: 

First, \textbf{data sparsity and the uncertainty of fine-grained information} are critical concerns. With coarse-grained sampling, vital short-term variations may not be adequately captured, resulting in significant uncertainty when inferring fine-grained states from coarse data. Second, \textbf{the transformation of spatio-temporal dependencies} is central to predicting from coarse to fine granularity. In traffic prediction, modeling spatio-temporal dependencies is crucial for accuracy \cite{zheng2023spatio,zhang2023mlpst}; thus, effectively mining spatio-temporal correlations from coarse data for fine-grained predictions poses a key challenge. Additionally, \textbf{the generalization capability of algorithmic} is an essential difficulty. Due to significant differences between coarse and fine-grained data, designing models that can train on coarse data and accurately predict fine-grained outcomes must overcome generalization challenges arising from inconsistent data scales. Finally, \textbf{the interpretability and computational resource limitations on traffic management systems} are factors that must be prioritized in practical applications, especially in scenarios like autonomous driving, where model transparency and computational constraints determine deployment feasibility. Balancing these concerns creates a fundamental trade-off between model complexity and operational feasibility, underscoring the need for architectures that are both accurate and resource-efficient.

To address these challenges, we formalize the fine-grained traffic prediction problem into two tasks—window-based and duration-based prediction—and propose the Spatio-Temporal Refinement Predictor (STRP) to solve them. STRP employs a tree-structured spatial convolution and inverse dilated convolution to achieve efficient and interpretable spatio-temporal modeling, robustly ensuring accurate recovery and prediction at low time complexity with minimal memory overhead. Specifically, we replace traditional graph convolution networks with a tree structure to hierarchically store the adjacency relationships among nodes, thereby reducing redundant computations. The computation process follows a bottom-up hierarchical order, where each layer’s convolution operation only involves the nodes at that level, avoiding the extensive unnecessary calculations found in full graph convolutions, thus significantly improving efficiency. At the same time, we introduce an inverse dilated convolution network for gradually restoring fine-grained traffic states. The core idea of inverse dilated convolution is to incrementally extrapolate from coarse-grained time series to fine-grained data. This process mimics recursive inference: deriving two nodes from one, then gradually expanding to four, and so forth. This layer-by-layer extrapolation not only facilitates a gradual refinement of time resolution but also effectively ensures the integrity of information transfer and temporal consistency.

Our main contributions are summarized as follows:
\begin{itemize}
    \item We formulate a new spatio-temporal data management problem: predicting fine-grained future traffic conditions from coarse-grained historical data, with the goal of enabling efficient and reliable fine-grained inference without incurring the cost of high-frequency data collection.
    \item We design an innovative spatial convolution structure, the tree convolution, which enables efficient and interpretable modeling of spatial dependencies.
    \item We propose inverse dilated convolution, which progressively and smoothly recovers fine-grained traffic states from coarse-grained data, ensuring high accuracy and spatio-temporal consistency in predictions.
    \item We evaluate the fine-grained prediction capabilities of our model against existing baseline models across multiple traffic benchmark datasets, demonstrating significant performance improvements.
\end{itemize}

\section{Problem Definition }
\begin{figure}[t]
\centerline{\includegraphics[width=0.5\textwidth]{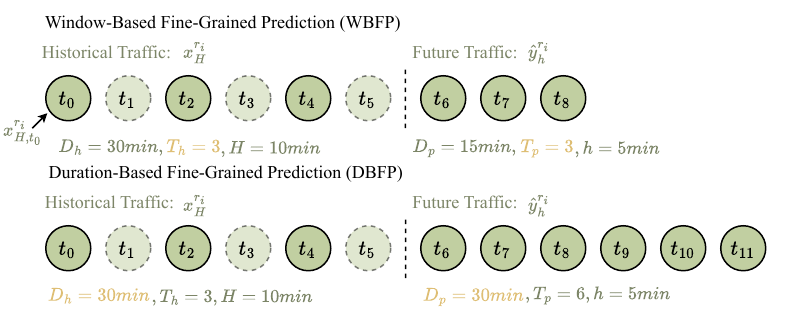}}
\caption{Examples are provided in the time dimension to illustrate the WBFP task and DBFP task.}
\label{fig: pro}
\end{figure}

We represent each road segment \( i \) as a node \( r_i \) in a tree structure. As shown in \textbf{Figure ~\ref{fig: pro}}, the traffic state of the node \( r_i \) at the \( t \)-th time step with a granularity coefficient \( H \) is denoted as \( x^{r_i}_{H,t} \). To represent the traffic state of the entire road network at the \( t \)-th time dimension with granularity coefficient \( H \), we define the matrix \( X_{H,t} = [x^{r_1}_{H,t}, x^{r_2}_{H,t}, \ldots, x^{r_N}_{H,t}] \), where \( N \) is the total number of road segments.

\noindent\textbf{\textit{Spatial Tree:}} To capture the spatial interdependencies, we construct a tree structure for each node \( r_i \). The child nodes of the tree are used to record other nodes that are adjacent to this node, with these relationships stored in an adjacency list \( \mathcal{N} \). This structure efficiently represents the spatial dependencies among nodes without significantly increasing storage costs.

In fine-grained traffic prediction, a fundamental challenge arises from temporal granularity mismatch—the fact that the same number of time steps may correspond to different actual time durations depending on the sampling resolution. This discrepancy leads to a semantic mismatch between the input and output sequences in the prediction task. To address this issue, we decompose the problem into two complementary sub-tasks: Window-Based Fine-Grained Prediction (WBFP) and Duration-Based Fine-Grained Prediction (DBFP). These two tasks are defined based on either a fixed number of time steps or a fixed temporal duration, respectively, and are designed to accommodate different real-world traffic scenarios, ranging from short-term, high-frequency forecasting to long-range granularity refinement.

\noindent\textbf{\textit{Problem 1 (WBFP)} :} The goal of WBFP is to perform short-term, fine-grained, accurate predictions, aiming to infer future fine-grained traffic states for the same time window from historical coarse-grained data. Specifically, given a historical time window \( T_h \) and a historical traffic state sequence \( X = \{X_{H,1}, X_{H,2}, \ldots, X_{H,T_h}\} \) with granularity coefficient \( H \), our objective is to predict the future fine-grained traffic states \( \hat{Y} = \{X_{h,(T_h+1)}, X_{h,(T_h+2)}, \ldots, X_{h,(T_h+T_p)}\} \), where \( T_p = T_h \) is the number of predicted time steps, and \( h \) is the fine-grained coefficient such that \( h \leq H \). This problem can be formalized as:
\begin{equation}
    \hat{Y} = f(X, \mathcal{N}; T_h, h), 
\end{equation}
where \( f \) denotes the prediction function, and \( \mathcal{N} \) represents the adjacency list.

\noindent\textbf{\textit{Problem 2 (DBFP)}:} DBFP infers future fine-grained traffic states with the same duration as the historical coarse-grained data, accounting for input-output granularity differences. Specifically, given a historical duration \( D_h \) and a corresponding historical traffic state sequence with granularity coefficient \( H \), where \( T_h = D_h/H \), we have \( X = \{X_{H,1}, X_{H,2}, \ldots, X_{H,T_h}\} \). Our objective is to predict \( \hat{Y} = \{X_{h,(T_h+1)}, X_{h,(T_h+2)}, \ldots, X_{h,(T_h+T_p)}\} \) with the same length, where \( T_p = T_h \cdot (H/h) \) represents the prediction time window, and \( h \leq H \). DBFP can be formalized as:

\begin{equation}
    \hat{Y} = f(X, \mathcal{N}; D_h, h). 
\end{equation}

\begin{table}[t]
  \centering
  \caption{Notations and corresponding descriptions.}
  \resizebox{\linewidth}{!}{
    \begin{tabular}{l|l}
    \toprule
    Notation & \multicolumn{1}{c}{Description} \\
    \midrule
    \( r_i, r_j \) & Road segments. \\
    \( N \) & Total number of road segments. \\
    \( H, h \) & Input granularity, output granularity (\( H \ge h \)). \\
    \( D_h, D_p \) & Historical traffic duration, predicted traffic duration. \\
    \multirow{2}[0]{*}{\( T_h, T_p \)} & Number of historical time windows, number of  \\
        & predicted time windows. \\
    \( \mathcal{N} \) & Adjacency list. \\ 
    \( \mathcal{N}(r_i) \) &  Row of node \( r_i \) in the adjacency list. \\
    \multirow{2}[0]{*}{\( x^{r_i}_{H,t} \)} & Traffic value of node \( r_i \) at granularity \( H \) in the \( t \)-th  \\
        & time slice. \\
    \( x^{r_i}_{H} \) & Input vector of node \( r_i \) at granularity \( H \). \\
    \multirow{2}[0]{*}{ \( X_{H,t} \)} & Traffic vector of all nodes at granularity \( H \) in the \( t \)-th \\
        &  time slice. \\
    \multirow{2}[0]{*}{\( X \)} & Traffic vector of all nodes at granularity \( H \) across all \\
        &  time slices. \\
    \( \hat{\mathbf{X}}_{H}^{t} \) & Traffic vector at granularity \( H \) after spatial aggregation. \\
    \( \mathbf{\hat{X}}_{h}^{t} \) & Expanded traffic vector at granularity \( h \). \\
    \bottomrule
    \end{tabular}%
    }
  \label{tab:notation}%
  \vspace{-0.25cm}
\end{table}%

For instance, with a 2-hour history at a sampling granularity of $H=30$ minutes ($T_h=4$ steps), setting the prediction granularity to $h=5$ minutes requires generating $T_p=24$ fine-grained steps within the same duration. This example illustrates the distinction between DBFP, which preserves duration, and WBFP, which preserves the number of time steps. We summarize the key notations in \textbf{Table \ref{tab:notation}}.

\section{Method}

\begin{figure*}[ht]
	\centering
		\includegraphics[width=1\linewidth]{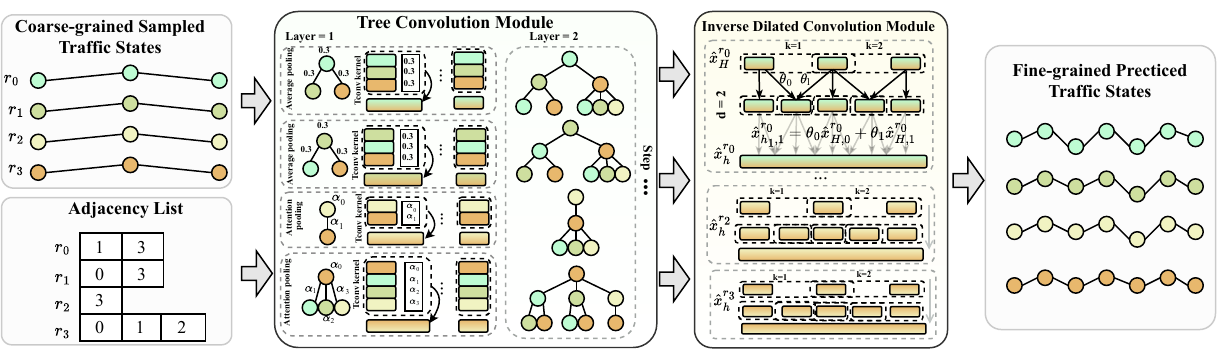}
		\caption{Overview of STRP, illustrated with DBFP task where \( D_h = D_p \). The Tree Convolution module shows two pooling methods, only one is actually used. We denote the STRP models with different pooling strategies as STRP$_{avg}$ and STRP$_{att}$.}
		\label{fig: STRP}
\end{figure*}

To address the fine-grained traffic prediction problem and its associated challenges, we propose the Spatio-Temporal Refinement Predictor (STRP). This model is designed to efficiently—and to a certain extent, interpretably—leverage coarse-grained sampled data for both window-based and duration-based fine-grained prediction tasks. As illustrated in \textbf{Figure~\ref{fig: STRP}}, STRP consists of two core components: the Tree Convolution Module, responsible for extracting spatial dependencies, and the Inverse Dilated Convolution Module, which performs temporal extrapolation for fine-grained prediction.

\begin{figure}[t]
\centerline{\includegraphics[width=0.5\textwidth]{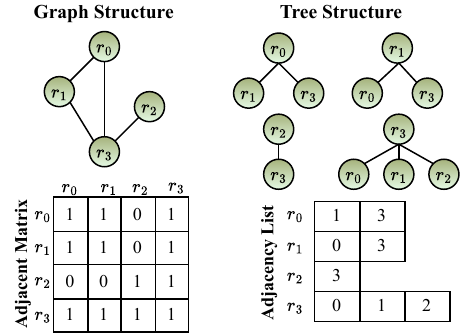}}
\caption{Comparison of graph structure and tree structure, where the adjacency list records the child nodes of each root.}
\label{fig: compare}
\end{figure}

\subsection{Tree Convolution Module}

Due to the inherent graph structure of traffic road networks, graph-based spatial aggregation methods such as Graph Convolutional Networks (GCNs) have become the most widely adopted approach for spatial modeling \cite{kipf2016semi,jiang2023spatio,lv2020temporal}. However, as the complexity of traffic data increases, spectral-based and matrix decomposition-based GCN methods have gradually revealed limitations in terms of high computational complexity and limited interpretability \cite{ju2024comprehensive, khoshraftar2024survey}. Some studies attempt to alleviate these issues through approximation techniques \cite{yu2017spatio} or sampling strategies, as in GraphSAGE \cite{hamilton2017inductive}. TreeCN introduces a tree-based structure into the spatial adjacency matrix to handle small-scale traffic scenarios \cite{lv2023treecn}, but its reliance on full-graph message propagation still imposes significant computational cost and limits interpretability.

Grounded in graph sparsification theory \cite{spielman2008graph}, we convert the dense road network graph $G(V, E)$ into a sparsified tree structure $T(V, E')$, where $|E'| \ll |E|$, retaining essential spatial dependencies while significantly reducing message-passing overhead. Building on this insight, we design a novel Tree Convolution Module. As illustrated in \textbf{Figure~\ref{fig: compare}}, the module adopts a pruning-inspired hierarchical structure to efficiently capture spatial correlations among nodes. Each layer aggregates information through a unique path from leaf nodes to the root, and the number of hops required for information propagation is bounded by the tree depth, thereby reducing the effective receptive field compared to multi-layer GCNs. During execution, the module applies a bottom-up recursive strategy, further reducing computational complexity while enhancing model interpretability.

\subsubsection{Adjacency List Construction}

To represent the spatial dependencies between nodes in the road network, we first constructed an adjacency list for each node \( r_i \). Given a set of nodes in the traffic network \( \mathcal{V} = \{r_1, r_2, \ldots, r_N\} \), the adjacency list \( \mathcal{N}(r_i) \) for each node \( r_i \) records its neighboring nodes, defined as:
\begin{equation}
    \mathcal{N}(r_i) = \{r_j \mid (r_i, r_j) \in \mathcal{E}\},
\end{equation}
where \( \mathcal{E} \) represents the set of edges in the road network, indicating the connections between nodes. This structure effectively captures the spatial correlations between nodes without introducing additional storage or computational overhead, ensuring the model's efficiency and scalability.

\subsubsection{Tree Convolution Operation} In graph-based spatial modeling, GCNs and Graph Attention Networks (GATs) are the two most widely adopted aggregation methods. We summarize and compare their respective formulations in \textbf{Section \ref{related:A}} as part of our related work discussion. Inspired by their design principles, we extend these aggregation mechanisms to a hierarchical tree structure tailored for our prediction task.

\begin{figure}[t]
\centerline{\includegraphics[width=0.4\textwidth]{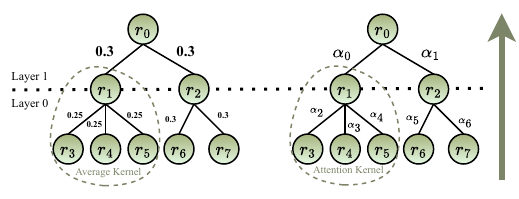}}
\caption{Tree convolutional hierarchical aggregation visualization. Each aggregation level in the tree provides a distinct semantic abstraction, progressively capturing spatial patterns from individual points to broader regions.}
\label{fig: explainability}
\end{figure}

In the Tree Convolution operation, nodes are organized into a pruned, layer-wise tree where information flows recursively from leaves to the root. Each row of the adjacency list specifies the child nodes of a root, and the hierarchy is expanded step by step as $\text{root} \rightarrow \text{child} \rightarrow \text{grandchild}$. This bottom-up flow enables structured and interpretable aggregation of node representations. Compared with full-graph convolutions, the unique propagation path reduces redundancy and improves transparency, as shown in \textbf{Figure~\ref{fig: explainability}}. To balance efficiency and expressiveness, we employ two strategies: average pooling for lightweight fusion and attention pooling for adaptive weighting of child contributions. The STRP framework can also accommodate other graph-based aggregation methods in future applications.


\textbf{Average Pooling:}Average pooling provides a simple yet structure-aware way to aggregate spatial features within the pruned tree. For feature updates of each node \( r_i \), average pooling uses the mean value of all its child nodes' features to update the parent node's feature. Specifically, at layer \( l \), the updated feature \( \mathbf{h}^{(l+1)}(r_i) \) of node \( r_i \) is given by:
\begin{equation}
\mathbf{h}^{(l+1)}(r_i) = \sigma\left( W^{(l)} \cdot \frac{1}{|\mathcal{N}(r_i)|} \sum_{r_j \in \mathcal{N}(r_i)} \mathbf{h}^{(l)}(r_j) + b^{(l)} \right),
\end{equation}
where \( W^{(l)} \) is the weight matrix at layer \( l \), \( b^{(l)} \) is the bias term, \( \sigma \) is a nonlinear activation function, and \( |\mathcal{N}(r_i)| \) denotes the number of child nodes of \( r_i \).Unlike conventional GCNs that operate over full neighborhoods, the tree-based average pooling limits aggregation to a unique parent-child path, reducing redundancy and propagation noise. This proves especially beneficial for coarse-to-fine prediction tasks, where over-smoothing from excessive neighbors can degrade fine-scale discriminability.

To quantify the information retention of tree-based aggregation, we consider the mutual information between the latent node representations from the tree structure $\mathbf{z}_i^{\text{tree}}$ and those from the full graph $\mathbf{z}_i^{\text{full}}$, measured as $I(\mathbf{z}_i^{\text{tree}} ; \mathbf{z}_i^{\text{full}}) = \mathcal{H}(\mathbf{z}_i^{\text{tree}}) - \mathcal{H}(\mathbf{z}_i^{\text{tree}} \mid \mathbf{z}_i^{\text{full}})$. A higher value indicates stronger alignment with full-graph representations, suggesting that the pruned structure still preserves essential spatial semantics. The unique message-passing path of trees reduces redundancy and improves gradient stability. However, the lack of overlapping neighborhoods may lead to missing higher-order dependencies, revealing a trade-off between interpretability and expressive capacity that warrants further investigation.

\textbf{Attention Pooling:} In contrast to average pooling, attention pooling dynamically adjusts the contribution of each child node based on its importance to the parent node’s representation. In the context of our tree structure, this process is restricted to layer-wise hierarchical paths, ensuring that message passing occurs exclusively along parent-child edges defined in the pruned tree. For each node $r_i$, its updated representation at layer $l+1$ is given by:
\begin{equation}
\mathbf{h}^{(l+1)}(r_i) = \sigma \left( W^{(l)} \cdot \mathcal{A} \left( \left\{ \mathbf{h}^{(l)}(r_j) \mid r_j \in \mathcal{N}(r_i) \right\} \right) + b^{(l)} \right),
\end{equation}
where  $\mathcal{A}(\cdot)$ represents the attention-based aggregation function. Unlike standard GAT, this formulation ensures that only structurally valid subtree paths contribute to the update, preventing information leakage from unrelated branches. The attention scores for each child $r_j$ are computed as:
\begin{equation}
    e_{ij} = \text{LeakyReLU} \left( \mathbf{a}^T \left[ W^{(l)} \mathbf{h}^{(l)}(r_i) \, \| \, W^{(l)} \mathbf{h}^{(l)}(r_j) \right] \right),
\end{equation}
\begin{equation}
    \alpha_{ij} = \frac{\exp(e_{ij})}{\sum_{r_k \in \mathcal{N}(r_i)} \exp(e_{ik})}.
\end{equation}

These attention weights $\alpha_{ij}$ are then used within the aggregation function $\mathcal{A} \left( \cdot \right) = \sum_{r_j} \alpha_{ij} \mathbf{h}^{(l)}(r_j)$, enabling selective focus on the most informative descendants. This tree-restricted attention scheme offers two advantages: (1) a reduced message passing space of $\mathcal{O}(N \log N)$ compared to $\mathcal{O}(N^2)$ in fully connected GATs, and (2) improved interpretability by enforcing a unique, layer-dependent path for information flow from leaf to root.

Average and attention pooling serve complementary purposes in our framework and are designed as interchangeable options rather than being used simultaneously. Average pooling offers high efficiency, making it suitable for real-time or resource-constrained settings, while attention pooling is better suited for complex spatial patterns due to its ability to adaptively weight child node importance. Both share a recursive bottom-up structure, but trade off between simplicity and expressive power.

\subsubsection{Multi-Step Tree Convolution Process}

To ensure that the hierarchical structure captures comprehensive spatial information across multiple levels, we perform multi-step tree convolutions. This allows the model to recursively aggregate spatial features from lower-level nodes up to the root across all layers. The final feature representation of node \( r_i \) at layer \( L \) is computed as:
\begin{equation}
\mathbf{h}^{(L)}(r_i) = \sigma \left( \sum_{l=1}^{L} W^{(l)} \cdot \mathcal{A}^{(l)} \left( \left\{ \mathbf{h}^{(l)}(r_j)   \right\} \right) + b^{(l)} \right),
\end{equation}
where $r_j \in  \mathcal{N}(r_i)$ and $\mathcal{A}^{(l)}(\cdot)$ denotes the layer-wise aggregation function defined as either average pooling or attention pooling. For average pooling, \( \alpha_{ij}^{(l)} = \frac{1}{|\mathcal{N}(r_i)|} \); for attention pooling, it is computed using the attention mechanism described earlier. 
This formulation decouples spatial aggregation from static neighborhood topology and instead anchors it within the recursive depth of the tree. This hierarchical formulation naturally supports efficient message propagation: the root node does not need to directly aggregate information from grandchildren, since child nodes have already integrated their own subtree representations in previous layers. This eliminates the need for cross-layer fusion while ensuring deep spatial context accumulation.

Each layer thus progressively captures broader semantic patterns—from local to regional scales—allowing global spatial information to be constructed through recursive application alone. Thanks to the logarithmic depth of the tree, this multi-step scheme remains computationally efficient while maintaining expressive capacity. The resulting node embeddings, enriched through this structured aggregation, are then passed to the temporal module for fine-grained sequence reconstruction. The overall process is detailed in \textbf{Algorithm~\ref{alg:treeconv}}.

\begin{algorithm}[t]
    \caption{Tree-Based Spatial Convolution}
    \label{alg:treeconv}
    \begin{algorithmic}[1]
        \STATE \textbf{Input:} Node features $x$, adjacency list $\mathcal{N}$, pooling type $\text{pool\_type} \in \{\text{avg}, \text{attention}\}$, number of layers $L$
        \STATE \textbf{Output:} Coarse-grained spatio-temporal representation $\mathbf{\hat{X}}_{H}^{t}$
        \FOR{each layer $l$ from $1$ to $L$}
            \FOR{each node $r_i$ in $x$}
                \STATE $\text{neighbors} \gets \mathcal{N}(r_i)$
                \IF{$\text{pool\_type} == \text{avg}$}
                    \STATE $\hat{x}_i \gets \sigma\left( \frac{1}{|\text{neighbors}|} \sum_{r_j \in \text{neighbors}} W^{(l)} x_j + b^{(l)} \right)$
                \ELSIF{$\text{pool\_type} == \text{attention}$}
                    \STATE Compute $e_{ij}$ for each $r_j$ via attention mechanism
                    \STATE $\alpha_{ij} \gets \text{softmax}(e_{ij})$
                    \STATE $\hat{x}_i \gets \sigma\left( \sum_{r_j \in \text{neighbors}} \alpha_{ij} W^{(l)} x_j + b^{(l)} \right)$
                \ENDIF
            \ENDFOR
        \ENDFOR
        \STATE $\mathbf{\hat{X}}_{H}^{t} \gets \hat{x}_H^t$
        \RETURN $\mathbf{\hat{X}}_{H}^{t}$
    \end{algorithmic}
\end{algorithm}

\subsection{Inverse Dilated Convolution}

\begin{figure}[t]
\centerline{\includegraphics[width=0.5\textwidth]{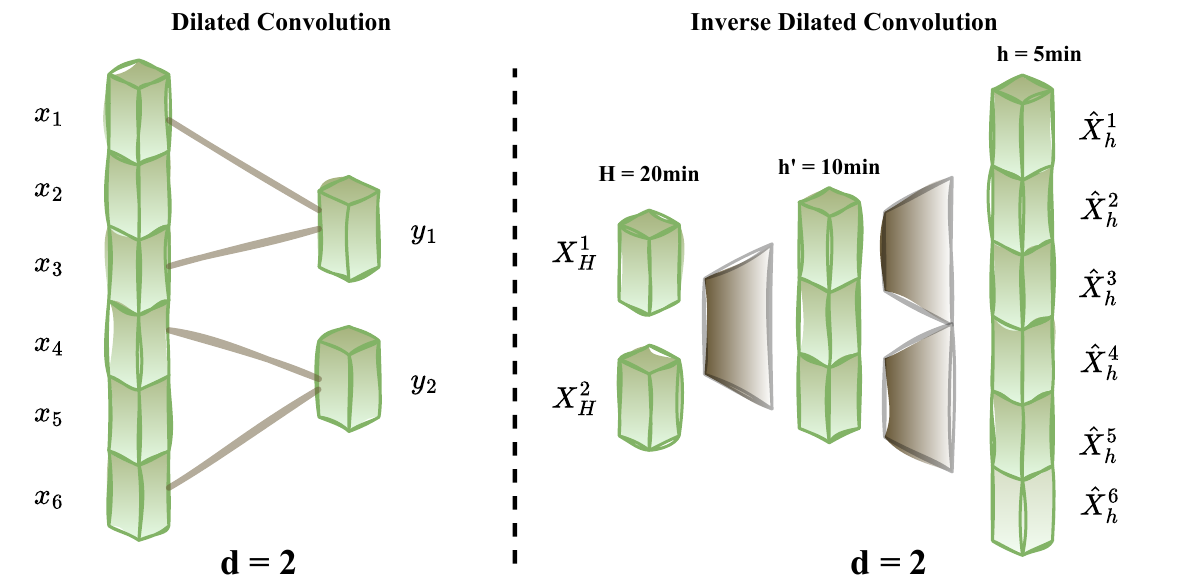}}
\caption{Design overview comparison of DConv and IDConv with $d=2$.}
\label{fig: IDConv}
\end{figure}

The Inverse Dilated Convolution (IDConv) module is designed to restore fine-grained temporal resolution from coarse-grained sequences, addressing the temporal granularity mismatch problem introduced earlier. As illustrated in \textbf{Figure~\ref{fig: IDConv}}, unlike standard dilated convolutions \cite{wu2019graph} that progressively expand the receptive field by skipping intermediate time steps, IDConv operates in the reverse direction: it gradually fills finer time slices by referencing coarser temporal signals. This process can be regarded as a hierarchical temporal decoder that unfolds coarse patterns into fine-grained sequences while maintaining temporal consistency and contextual coherence. Such layer-by-layer unfolding not only preserves the causal order of time series but also limits the accumulation of single-step prediction errors through recursive insertion of intermediate states.

\subsubsection{Design of the Diffusion Convolution Kernel}

IDConv introduces a diffusion convolution kernel in the temporal dimension to progressively refine coarse-grained slices. This process resembles unfolding a rough time series into a more detailed sequence. Unlike conventional temporal convolutions that compress or abstract sequential information, this module acts as a temporal “decoder”—gradually expanding the sequence toward finer resolution. By incorporating nonlinear transformations and parameter sharing at each layer, IDConv preserves temporal correlations while suppressing error amplification, thereby stabilizing upsampling results in both statistical distribution and temporal continuity.

From the Tree Convolution Module, we obtain the coarse-grained historical traffic sequence $\mathbf{\hat{X}}_{H}^{t}$, where $H$ denotes the coarse interval and $t$ the time step. IDConv then refines the temporal granularity step by step using the diffusion kernel. For each time step $t$, the output $\mathbf{\hat{X}}_{h}^{t}$ is computed as:
\begin{equation}
\mathbf{\hat{X}}^{t}_{h} = \sum_{k=0}^{K} \theta_k \mathbf{\hat{X}}_{H}^{t - k \cdot d},
\end{equation}
where $K$ is the kernel size, $d$ the diffusion stride (dilation gap), and $\theta_k$ the parameter of the $k$-th kernel. A smaller stride $d$ allows denser interpolation between coarse inputs, enabling more accurate restoration of intermediate states. The learnable parameters $\theta_k$ serve as interpolation weights, distributing coarse-level semantics across the finer domain. This mechanism is especially effective in our setting, as it recursively constructs denser, temporally aligned sequences without requiring fine-grained ground truth at every step.

The term \textit{inverse} emphasizes its fundamental contrast to standard dilated convolutions. While dilation skips steps to capture coarse patterns, the inverse variant upsamples sequences by recursively inserting intermediate states, enabling fine-grained recovery with causal consistency. Compared with conventional Temporal Convolutional Networks (TCNs) \cite{bai2018empirical,yu2017spatio}, which expand receptive fields through dilated sampling, IDConv focuses on recursive interpolation and refinement to address cross-granularity forecasting. Thus, the two differ essentially in both objectives and mechanisms.

\subsubsection{Fine-Grained Restoration}

During restoration, IDConv recursively extrapolates coarse sequences into high-resolution predictions. The process begins with coarse temporal points and progressively unfolds into finer steps, reconstructing future fine-grained states via hierarchical refinement. We denote the first intermediate prediction as $\mathbf{Y}^{(1)} = \mathbf{\widetilde{X}}_{h}^{t}$, initialized from Tree Convolution outputs. At the $l$-th layer, the fine-grained state $\mathbf{Y}^{(l)}$ is updated by applying IDConv to the previous iteration rather than static inputs:

\begin{equation}
\mathbf{Y}^{(l+1)} = \sigma \left( \sum_{t=1}^{T} \theta^{(l)} \ast \mathbf{Y}^{(l)} \right),
\end{equation}

where $\ast$ denotes temporal convolution, $\theta^{(l)}$ is the diffusion kernel at layer $l$, and $\sigma$ is a nonlinear activation. Importantly, this recursive formulation ensures $\mathbf{Y}^{(l+1)}$ explicitly depends on $\mathbf{Y}^{(l)}$, thus preserving temporal consistency across refinement steps and avoiding excessive error amplification during upsampling. Through multiple layers, the coarse sequence $\mathbf{\hat{X}}_H^t$ is refined into the final fine-grained prediction $\mathbf{\hat{Y}}$.

This hierarchical refinement process preserves temporal continuity while gradually increasing resolution, instead of producing fine-grained outputs in a single pass. It also leverages coarse semantic context at each stage, reinforcing cross-scale consistency. The detailed recursive procedure is outlined in \textbf{Algorithm~\ref{alg:idconv}}.

\begin{algorithm}[t]
    \caption{Recursive Fine-Grained Restoration via IDConv}
    \label{alg:idconv}
    \begin{algorithmic}[1]
        \STATE \textbf{Input:} Coarse-grained $\mathbf{\hat{X}}_{H}^{t}$, kernel parameters $\theta$, target length $T_p$, dilation stride $d = H/h$, number of layers $L$
        \STATE \textbf{Output:} Fine-grained sequence $\hat{Y}$
        \STATE Initialize $\mathbf{Y}^{(1)} \gets x$ \COMMENT{Initial coarse features}
        \FOR{$l \gets 1$ \TO $L$}
            \STATE Initialize $\mathbf{Y}^{(l+1)} \gets []$
            \FOR{$t \gets 0$ \TO $T_p - 1$}
                \STATE $\text{expanded\_features} \gets []$
                \FOR{$k \gets 0$ \TO $K - 1$}
                    \STATE $\text{step} \gets t - k \cdot d$
                    \IF{$\text{step} \geq 0$}
                        \STATE $\text{expanded\_features.append}(\theta_k \cdot \mathbf{Y}^{(l)}[\text{step}])$
                    \ENDIF
                \ENDFOR
                \STATE $\mathbf{Y}^{(l+1)}_t \gets \sigma\left(\sum \text{expanded\_features}\right)$
            \ENDFOR
        \ENDFOR
        \STATE $\mathbf{\hat{Y}} \gets \mathbf{Y}^{(L+1)}$
        \RETURN $\mathbf{\hat{Y}}$
    \end{algorithmic}
\end{algorithm}

\subsection{Training Strategy}

To optimize the STRP model for fine-grained traffic prediction, we adopt the Mean Squared Error (MSE) loss, which directly penalizes deviations between predicted and true values at each node and each fine-grained time step. This choice is motivated by its simplicity and effectiveness in continuous regression tasks, particularly in capturing magnitude-sensitive deviations. The MSE loss is computed as:
\begin{equation}
    \mathcal{L}_{MSE} = \frac{1}{T_p} \sum^{T_p}_{t = 1} \frac{1}{N} \sum^{N}_{i = 1} \left( y^{r_i}_t - \widehat{y}^{r_i}_t \right)^2,
\end{equation}
where \( y^{r_i}_t \) denotes the ground truth traffic value for node \( r_i \) at fine-grained time step \( t \), and \( \widehat{y}^{r_i}_t \) is the corresponding predicted value. This formulation ensures that model learning is directly aligned with the objective of restoring temporal resolution from coarse-grained observations. The full end-to-end training procedure of STRP is summarized in \textbf{Algorithm~\ref{alg:training}}.

\begin{algorithm}[t]
    \caption{Training Loop for STRP Model}
    \label{alg:training}
    \begin{algorithmic}[1]
        \STATE \textbf{Input:} Traffic sequence $X$, adjacency list $\mathcal{N}$, pooling type $\text{pool\_type}$, temporal settings $(T_h, T_p, h, H)$, max epochs $\text{max\_epochs}$, batch size $\text{batch\_size}$
        \FOR{$\text{epoch} \gets 1$ \TO $\text{max\_epochs}$}
            \STATE $\text{batches} \gets \text{create\_batches}(X, \text{batch\_size})$
            \FORALL{$\text{batch}$ in $\text{batches}$}
                \STATE $\hat{X}_H \gets \text{TreeConvolution}(\text{batch}, \mathcal{N}, \text{pool\_type})$
                \STATE $\hat{Y} \gets \text{InverseDilatedConvolution}(\hat{X}_H, T_p, h, H)$
                \STATE $\mathcal{L}_{MSE} \gets \text{compute\_loss}(\hat{Y}, Y)$
                \STATE $\text{update\_parameters}(\mathcal{L}_{MSE})$
            \ENDFOR
        \ENDFOR
    \end{algorithmic}
\end{algorithm}

\section{Experiments}

We conducted experiments on six widely used traffic benchmark datasets to validate the fine-grained future traffic state prediction problem and demonstrate the effectiveness of our STRP model. We compared STRP's performance with 12 existing traffic prediction models across two fine-grained prediction tasks.

\subsection{Datasets Description}

The datasets used in the experiments include the METR-LA dataset, collected from loop detectors on Los Angeles freeways, and five benchmark datasets (PeMS-Bay, PeMS03, PeMS04, PeMS07, and PeMS08) collected by the California Department of Transportation Performance Measurement System (PeMS) \cite{luo2023stg4traffic,li2023dynamic}. All datasets have a unified recording interval of 5 minutes. These datasets cover traffic information across different temporal and spatial ranges. Detailed statistics of these datasets are provided in \textbf{Table~\ref{tab:datasets}}.

\begin{table}[t]
  \centering
  \caption{Statistics of all datasets.}
    \begin{tabular}{cccc}
    \toprule
    Dataset & METR-LA & PeMS-Bay & PeMS03 \\
    \midrule
    Start Time & 2012/3/1 & 2017/1/1 & 2018/9/1 \\
    End Time & 2012/6/30 & 2017/5/31 & 2018/11/30 \\
    Time Interval & 5 min & 5 min & 5 min \\
    \# Object & 207 & 325 & 358 \\
    \# Timestamps & 34272 & 52116 & 26208 \\
    \midrule
    Dataset & PeMS04 & PeMS07 & PeMS08 \\
    \midrule
    Start Time & 2018/1/1 & 2017/5/1 & 2016/7/1 \\
    End Time & 2018/2/28 & 2017/8/31 & 2016/8/31 \\
    Time Interval & 5 min & 5 min & 5 min \\
    \# Object & 307 & 883 & 170 \\
    \# Timestamps & 16992 & 28224 & 17856 \\
    \bottomrule
    \end{tabular}%
  \label{tab:datasets}%
\end{table}%

To simulate traffic prediction scenarios with varying granularities, we adjusted the original data's sampling intervals to 10, 20, and 30 minutes. In all experiments, we used a historical input time span of 6 hours, resulting in time windows of 36, 18, and 12 for the respective sampling intervals. Each dataset was split into training, validation, and test sets in a 6:2:2 ratio based on chronological order.

\subsection{Evaluation Metrics}

To comprehensively assess the accuracy of fine-grained traffic prediction, we employ three standard evaluation metrics: Mean Absolute Error (MAE), Root Mean Squared Error (RMSE), and Mean Absolute Percentage Error (MAPE). These metrics respectively quantify the average absolute error, the squared error with a greater penalty for large deviations, and the relative error scaled by ground truth values. All are widely used in spatio-temporal prediction tasks, and lower values indicate better performance. The formal definitions of these metrics are given below:

\noindent\textbf{MAE}: Measures the average absolute difference between the predicted values and the actual values.
\begin{equation}
MAE = \frac{1}{T_p}\sum^{T_p}_{t = 1}\frac{1}{N}\sum^{N}_{i = 1}\Big|y^{r_i}_t - \widehat{y}^{r_i}_t \Big|.
\end{equation}
\noindent\textbf{RMSE}: Reflects the square root of the average squared differences between predicted values and actual values, emphasizing the impact of larger errors.
\begin{equation}
RMSE = \sqrt{\frac{1}{T_p}\sum^{T_p}_{t = 1}\frac{1}{N}\sum^{N}_{i = 1}\Big(y^{r_i}_t - \widehat{y}^{r_i}_t \Big)^2}.
\end{equation}
\noindent\textbf{MAPE}: Measures the percentage difference between the predicted and actual values, reflecting the size of the relative error.
\begin{equation}
MAPE = \frac{1}{T_p}\sum^{T_p}_{t = 1}\frac{1}{N}\sum^{N}_{i = 1}\Big|\frac{y^{r_i}_t - \widehat{y}^{r_i}_t}{y^{r_i}_t} \Big|.
\end{equation}
\begin{table*}[ht]
  \centering
  \caption{Results on the WBFP task, with \( T_h = T_p = 18 \).}
  \resizebox{\linewidth}{!}{
  \setlength{\tabcolsep}{2pt}
    \begin{tabular}{c|c|ccccccccccccccc}
    \toprule
    Datasets & Metrics & DCRNN & STGCN & ASTGCN & GWNet & STSGCN & MTGNN & AGCRN & STGODE & STID  & STAEf. & Traff. & MegaCRN & TimeMixer & \boldmath{}\textbf{STRP$_{avg}$}\unboldmath{} & \boldmath{}\textbf{STRP$_{att}$}\unboldmath{} \\
    \midrule
    \multirow{3}[2]{*}{METR-LA} & MAE   & 9.09  & 9.45  & 16.01  & 8.83  & 10.86  & 8.83  & 9.38  & 8.92  & 9.27  & 8.70  & 8.82  & 8.27  & \cellcolor[rgb]{ .906,  .902,  .902}7.90  & \cellcolor[rgb]{ .776,  .878,  .706}\textbf{5.68 } & \cellcolor[rgb]{ .663,  .816,  .557}\textbf{5.22 } \\
          & RMSE  & 12.13  & 12.94  & 20.89  & 11.61  & 17.18  & 11.68  & 12.51  & 11.96  & 12.61  & 11.56  & 11.75  & 11.13  & \cellcolor[rgb]{ .906,  .902,  .902}10.00  & \cellcolor[rgb]{ .776,  .878,  .706}\textbf{9.85 } & \cellcolor[rgb]{ .663,  .816,  .557}\textbf{8.97 } \\
          & MAPE  & 24.28\% & 25.34\% & 30.63\% & 22.95\% & 26.80\% & 22.81\% & 25.11\% & 24.66\% & 25.81\% & 22.68\% & 23.49\% & \cellcolor[rgb]{ .906,  .902,  .902}21.42\% & 21.48\% & \cellcolor[rgb]{ .776,  .878,  .706}\textbf{15.53\%} & \cellcolor[rgb]{ .663,  .816,  .557}\textbf{14.44\%} \\
    \midrule
    \multirow{3}[2]{*}{PeMS-Bay} & MAE   & 6.19  & 6.11  & 6.82  & 5.84  & 6.46  & 5.93  & 6.11  & 5.60  & 5.95  & 5.84  & 5.93  & 5.75  & \cellcolor[rgb]{ .906,  .902,  .902}5.17  & \cellcolor[rgb]{ .776,  .878,  .706}\textbf{3.34 } & \cellcolor[rgb]{ .663,  .816,  .557}\textbf{2.80 } \\
          & RMSE  & 10.50  & 10.53  & 11.14  & 9.75  & 10.71  & 9.93  & 10.25  & 9.72  & 10.03  & 9.86  & 9.91  & 9.68  & \cellcolor[rgb]{ .906,  .902,  .902}9.16  & \cellcolor[rgb]{ .776,  .878,  .706}\textbf{6.53 } & \cellcolor[rgb]{ .663,  .816,  .557}\textbf{5.47 } \\
          & MAPE  & 16.62\% & 16.62\% & 18.45\% & 15.65\% & 17.42\% & 15.88\% & 16.79\% & 15.69\% & 16.24\% & 15.59\% & 16.15\% & 15.30\% & \cellcolor[rgb]{ .906,  .902,  .902}14.27\% & \cellcolor[rgb]{ .776,  .878,  .706}\textbf{8.02\%} & \cellcolor[rgb]{ .663,  .816,  .557}\textbf{6.02\%} \\
    \midrule
    \multirow{3}[2]{*}{PeMS03} & MAE   & 35.86  & 34.50  & 34.89  & 39.15  & 34.48  & 32.22  & 31.68  & 32.55  & 31.87  & 31.74  & 34.50  & 30.19  & \cellcolor[rgb]{ .906,  .902,  .902}28.86  & \cellcolor[rgb]{ .776,  .878,  .706}\textbf{13.10 } & \cellcolor[rgb]{ .663,  .816,  .557}\textbf{12.28 } \\
          & RMSE  & 54.18  & 53.84  & 29..66 & 58.88  & 52.21  & 48.58  & 50.92  & 49.76  & 52.36  & 48.11  & 53.82  & 46.33  & \cellcolor[rgb]{ .906,  .902,  .902}41.61  & \cellcolor[rgb]{ .776,  .878,  .706}\textbf{20.21 } & \cellcolor[rgb]{ .663,  .816,  .557}\textbf{18.53 } \\
          & MAPE  & 42.10\% & 38.18\% & 43.19\% & 42.99\% & 37.36\% & 34.37\% & 35.29\% & 37.16\% & 36.42\% & 34.17\% & 39.23\% & 32.47\% & \cellcolor[rgb]{ .906,  .902,  .902}32.37\% & \cellcolor[rgb]{ .776,  .878,  .706}\textbf{26.61\%} & \cellcolor[rgb]{ .663,  .816,  .557}\textbf{26.37\%} \\
    \midrule
    \multirow{3}[2]{*}{PeMS04} & MAE   & 46.54  & 42.78  & 43.21  & 47.96  & 39.93  & 41.52  & 37.37  & 39.27  & 35.57  & 40.89  & 43.07  & 40.26  & \cellcolor[rgb]{ .906,  .902,  .902}36.20  & \cellcolor[rgb]{ .776,  .878,  .706}\textbf{18.95 } & \cellcolor[rgb]{ .663,  .816,  .557}\textbf{16.51 } \\
          & RMSE  & 66.11  & 61.65  & 61.08  & 68.85  & 58.41  & 58.11  & 55.95  & 56.92  & 53.86  & 57.70  & 60.80  & 56.66  & \cellcolor[rgb]{ .906,  .902,  .902}53.62  & \cellcolor[rgb]{ .776,  .878,  .706}\textbf{27.88 } & \cellcolor[rgb]{ .663,  .816,  .557}\textbf{24.35 } \\
          & MAPE  & 39.71\% & 33.84\% & 38.41\% & 40.10\% & 32.24\% & 32.32\% & 30.08\% & 31.94\% & 28.53\% & 31.75\% & 38.16\% & 31.15\% & \cellcolor[rgb]{ .906,  .902,  .902}29.06\% & \cellcolor[rgb]{ .776,  .878,  .706}\textbf{22.29\%} & \cellcolor[rgb]{ .663,  .816,  .557}\textbf{18.86\%} \\
    \midrule
    \multirow{3}[2]{*}{PeMS07} & MAE   & 51.85  & 52.01  & 57.49  & 55.03  & 49.72  & 48.58  & 43.63  & 47.12  & 41.37  & 47.86  & 49.02  & 45.51  & \cellcolor[rgb]{ .906,  .902,  .902}42.16  & \cellcolor[rgb]{ .776,  .878,  .706}\textbf{39.30 } & \cellcolor[rgb]{ .663,  .816,  .557}\textbf{30.24 } \\
          & RMSE  & 76.11  & 76.51  & 83.99  & 84.40  & 77.00  & 69.05  & 69.29  & 74.06  & 65.75  & 68.38  & 78.31  & 66.85  & \cellcolor[rgb]{ .906,  .902,  .902}66.27  & \cellcolor[rgb]{ .776,  .878,  .706}\textbf{58.28 } & \cellcolor[rgb]{ .663,  .816,  .557}\textbf{40.90 } \\
          & MAPE  & 27.45\% & 26.09\% & 32.77\% & 28.54\% & 24.04\% & 23.49\% & 21.12\% & 23.97\% & 20.01\% & 23.35\% & 25.46\% & 22.05\% & \cellcolor[rgb]{ .906,  .902,  .902}20.30\% & \cellcolor[rgb]{ .776,  .878,  .706}\textbf{18.46\%} & \cellcolor[rgb]{ .663,  .816,  .557}\textbf{16.30\%} \\
    \midrule
    \multirow{3}[2]{*}{PeMS08} & MAE   & 35.19  & 35.50  & 36.67  & 37.69  & 33.75  & 34.46  & 31.42  & 33.12  & 28.30  & 33.94  & 34.00  & 33.41  & \cellcolor[rgb]{ .906,  .902,  .902}29.99  & \cellcolor[rgb]{ .776,  .878,  .706}\textbf{14.63 } & \cellcolor[rgb]{ .663,  .816,  .557}\textbf{13.66 } \\
          & RMSE  & 53.59  & 53.69  & 54.32  & 59.90  & 51.70  & 50.60  & 48.65  & 50.10  & 45.88  & 50.24  & 52.55  & 49.33  & \cellcolor[rgb]{ .906,  .902,  .902}46.80  & \cellcolor[rgb]{ .776,  .878,  .706}\textbf{22.19 } & \cellcolor[rgb]{ .663,  .816,  .557}\textbf{20.34 } \\
          & MAPE  & 38.36\% & 38.19\% & 43.82\% & 42.48\% & 36.71\% & 36.48\% & 33.80\% & 35.58\% & 30.23\% & 35.83\% & 37.85\% & 35.16\% & \cellcolor[rgb]{ .906,  .902,  .902}32.39\% & \cellcolor[rgb]{ .776,  .878,  .706}\textbf{20.43\%} & \cellcolor[rgb]{ .663,  .816,  .557}\textbf{19.34\%} \\
    \bottomrule
    \end{tabular}%
  \label{tab:main_WBFP}%
  }
\end{table*}%
\subsection{Baseline Models}  

We compare STRP with 13 state-of-the-art traffic prediction models spanning three architectural paradigms:

\begin{itemize}
    \item \textbf{Graph-based models:} DCRNN~\cite{li2017diffusion}, STGCN~\cite{yu2017spatio}, ASTGCN~\cite{guo2019attention}, GWNet~\cite{wu2019graph}, STSGCN~\cite{song2020spatial}, MTGNN~\cite{yu2017spatio}, AGCRN \cite{bai2020adaptive}, STGODE~\cite{fang2021spatial}, and MegaCRN~\cite{jiang2023spatio}. These methods leverage graph structures to capture spatial dependencies in traffic networks.

    \item \textbf{Transformer-based model:} STAEformer (STAEf.)~\cite{liu2023spatio} and Trafformer (Traff.)\cite{jin2023trafformer}, which introduces attention mechanisms for modeling long-range spatio-temporal interactions.

    \item \textbf{MLP-based models:} STID~\cite{shao2022spatial} and TimeMixer~\cite{wangtimemixer}, which adopt lightweight architectures for efficient temporal modeling without relying on explicit spatial graphs.
\end{itemize}

Since this work is the first to formalize fine-grained prediction, existing models are not designed for this setting. To ensure a fair comparison under our fine-grained prediction setting, we uniformly adjust the output window length and target temporal granularity across all baselines. 

\subsection{Experimental Setup}

All experiments were conducted on a high-performance computing platform equipped with an Intel® Xeon® Max 9462 CPU (2.70 GHz) and six NVIDIA A100 80GB SXM GPUs. We used the Adam stochastic gradient descent algorithm to train the model, with an initial learning rate set to $lr = 1 \times 10^{-4}$, a batch size of 64, and a maximum of 1,000 training iterations. To prevent overfitting, we employed an early stopping strategy based on the performance on the validation set. Additionally, starting from the 20th training epoch, the learning rate was halved every 10 epochs to facilitate more stable model training.


\subsection{Main Results}

In the main experiments, we set the time granularity to 20 minutes and defined the input time window as 18 time steps, conducting two prediction tasks: WBFP and DBFP. The goal of WBFP is to predict traffic data for the next 1.5 hours (18 five-minute intervals), while DBFP aims to predict traffic data for the next 6 hours (72 five-minute intervals). The best results are highlighted with a dark green background, the second-best results with light green, and the third-best results with gray.

\textbf{Table~\ref{tab:main_WBFP}} presents the comparison results of the two STRP variants against baseline models in the WBFP task. The results show that existing traffic prediction models have clear shortcomings in fine-grained inference. For instance, ASTGCN performs well in traditional prediction but suffers larger errors here, likely due to its reliance on cyclic patterns such as daily and weekly correlations that are difficult to model from coarse-grained inputs. Trafformer achieves reasonable results but still falls short of STRP. STRP${avg}$ and STRP${att}$ achieved the best overall performance. Compared to TimeMixer, the strongest baseline, their average MAE across six datasets was reduced by about 70\% and 92\%, respectively. These findings confirm the necessity of specialized fine-grained models and demonstrate the effectiveness of STRP’s tree-structured spatial convolution and inverse dilated convolution.

\textbf{Table~\ref{tab:DBFP}} displays the performance comparison results of the two variants of the STRP model against baseline models in the DBFP task. The results show that existing traffic prediction models significantly underperform in the full-scale fine-grained prediction tasks over longer horizons, particularly on the PeMS03 and PeMS04 datasets, where some models had MAE errors as high as 107. This further reveals the unique challenges faced in full-scale fine-grained prediction: it is necessary to make predictions at a higher "resolution" time scale while maintaining accuracy over a longer time frame. In contrast, the STRP model exhibited outstanding performance in this task, especially STRP$_{att}$, which achieved the best results across all datasets and evaluation metrics, demonstrating high adaptability to fine-grained and long-term traffic prediction tasks. Although STRP$_{avg}$ was slightly inferior to STRP$_{att}$, it still significantly outperformed other baseline models. Compared to traditional spatio-temporal prediction models (such as DCRNN, STGCN, GWNet, etc.), STRP$_{att}$ achieved a substantial reduction in error metrics, indicating the design effectiveness and advantages of the STRP model. The performance of Trafformer was less stable across datasets, further underscoring the robustness of STRP. Among the baseline models, TimeMixer, AGCRN, and MegaCRN performed relatively well; however, their performance still showed a significant gap compared to the two variants of STRP, further validating the leading advantage of the STRP model in full-scale fine-grained prediction tasks.

Overall, the experimental results indicate that traditional traffic prediction models face unique difficulties and limitations in fine-grained prediction, while the STRP model demonstrates significant advantages in fine-grained traffic prediction tasks, achieving more accurate traffic state predictions at higher time resolutions and over longer time ranges.

\begin{table*}[t]
  \centering
  \caption{Results on the DBFP task, with \( D_h = D_p = 6h \).}
  \resizebox{\linewidth}{!}{
  \setlength{\tabcolsep}{2pt}
      \begin{tabular}{c|c|ccccccccccccccc}
    \toprule
    Datasets & Metrics & DCRNN & STGCN & ASTGCN & GWNet & STSGCN & MTGNN & AGCRN & STGODE & STID  & STAEf. & Traff. & MegaCRN & TimeMixer & \boldmath{}\textbf{STRP$_{avg}$}\unboldmath{} & \boldmath{}\textbf{STRP$_{att}$}\unboldmath{} \\
    \midrule
    \multirow{3}[2]{*}{METR-LA} & MAE   & 11.12  & 11.57  & 19.60  & 10.80  & 13.29  & 10.80  & 11.49  & 10.91  & 11.34  & 10.64  & 10.79  & 10.12  & \cellcolor[rgb]{ .906,  .902,  .902}9.67  & \cellcolor[rgb]{ .886,  .937,  .855}\textbf{7.63 } & \cellcolor[rgb]{ .663,  .816,  .557}\textbf{7.02 } \\
          & RMSE  & 13.17  & 14.05  & 22.70  & 12.61  & 18.66  & 12.68  & 13.59  & 12.99  & 13.69  & 12.56  & 12.59  & 12.09  & \cellcolor[rgb]{ .906,  .902,  .902}10.86  & \cellcolor[rgb]{ .886,  .937,  .855}\textbf{9.36 } & \cellcolor[rgb]{ .663,  .816,  .557}\textbf{8.43 } \\
          & MAPE  & 30.05\% & 31.37\% & 37.92\% & 28.41\% & 33.18\% & 28.24\% & 31.08\% & 30.53\% & 31.96\% & 28.08\% & 29.08\% & \cellcolor[rgb]{ .906,  .902,  .902}26.51\% & 26.60\% & \cellcolor[rgb]{ .886,  .937,  .855}\textbf{21.52\%} & \cellcolor[rgb]{ .663,  .816,  .557}\textbf{20.74\%} \\
    \midrule
    \multirow{3}[2]{*}{PeMS-Bay} & MAE   & 6.21  & 6.12  & 6.84  & 5.85  & 6.48  & 5.94  & 6.12  & 5.62  & 5.97  & 5.85  & 5.94  & 5.76  & \cellcolor[rgb]{ .906,  .902,  .902}5.18  & \cellcolor[rgb]{ .886,  .937,  .855}\textbf{4.16 } & \cellcolor[rgb]{ .663,  .816,  .557}\textbf{3.08 } \\
          & RMSE  & 10.52  & 10.55  & 11.16  & 9.77  & 10.73  & 9.95  & 10.27  & 9.74  & 10.05  & 9.88  & 9.89  & 9.70  & \cellcolor[rgb]{ .906,  .902,  .902}9.18  & \cellcolor[rgb]{ .886,  .937,  .855}\textbf{7.96 } & \cellcolor[rgb]{ .663,  .816,  .557}\textbf{6.18 } \\
          & MAPE  & 16.68\% & 16.68\% & 18.52\% & 15.70\% & 17.48\% & 15.93\% & 16.85\% & 15.74\% & 16.30\% & 15.65\% & 16.21\% & 15.35\% & \cellcolor[rgb]{ .906,  .902,  .902}14.32\% & \cellcolor[rgb]{ .886,  .937,  .855}\textbf{10.19\%} & \cellcolor[rgb]{ .663,  .816,  .557}\textbf{8.81\%} \\
    \midrule
    \multirow{3}[2]{*}{PeMS03} & MAE   & 98.33  & 94.60  & 95.68  & 107.37  & 94.55  & 88.36  & 86.87  & 89.25  & 87.38  & 87.05  & 93.35  & 82.78  & \cellcolor[rgb]{ .906,  .902,  .902}79.13  & \cellcolor[rgb]{ .886,  .937,  .855}\textbf{27.71 } & \cellcolor[rgb]{ .663,  .816,  .557}\textbf{21.30 } \\
          & RMSE  & 133.35  & 132.52  & 29.66  & 144.93  & 128.51  & 119.59  & 125.35  & 122.49  & 128.87  & 118.43  & 124.48  & 114.05  & \cellcolor[rgb]{ .906,  .902,  .902}102.42  & \cellcolor[rgb]{ .886,  .937,  .855}\textbf{40.42 } & \cellcolor[rgb]{ .663,  .816,  .557}\textbf{32.41 } \\
          & MAPE  & 203.44\% & 184.50\% & 208.71\% & 207.74\% & 180.52\% & 166.10\% & 170.52\% & 179.55\% & 175.99\% & 165.14\% & 175.59\% & \cellcolor[rgb]{ .906,  .902,  .902}155.93\% & 156.42\% & \cellcolor[rgb]{ .886,  .937,  .855}\textbf{61.61\%} & \cellcolor[rgb]{ .663,  .816,  .557}\textbf{42.20\%} \\
    \midrule
    \multirow{3}[2]{*}{PeMS04} & MAE   & 81.35  & 74.77  & 75.53  & 83.83  & 69.79  & 72.57  & 65.31  & 68.64  & 62.16  & 71.48  & 73.52  & 70.37  & \cellcolor[rgb]{ .906,  .902,  .902}63.27  & \cellcolor[rgb]{ .886,  .937,  .855}\textbf{35.48 } & \cellcolor[rgb]{ .663,  .816,  .557}\textbf{27.96 } \\
          & RMSE  & 111.87  & 104.33  & 103.36  & 116.51  & 98.84  & 98.34  & 94.67  & 96.32  & 91.13  & 97.64  & 99.04  & 95.87  & \cellcolor[rgb]{ .906,  .902,  .902}90.74  & \cellcolor[rgb]{ .886,  .937,  .855}\textbf{51.07 } & \cellcolor[rgb]{ .663,  .816,  .557}\textbf{40.67 } \\
          & MAPE  & 107.55\% & 91.66\% & 104.03\% & 108.62\% & 87.32\% & 87.55\% & 81.48\% & 86.50\% & 77.27\% & 85.99\% & 96.06\% & 84.39\% & \cellcolor[rgb]{ .906,  .902,  .902}78.71\% & \cellcolor[rgb]{ .886,  .937,  .855}\textbf{55.14\%} & \cellcolor[rgb]{ .663,  .816,  .557}\textbf{44.30\%} \\
    \midrule
    \multirow{3}[2]{*}{PeMS07} & MAE   & 62.29  & 62.48  & 69.06  & 66.10  & 59.73  & 58.36  & 52.41  & 56.60  & 49.70  & 57.49  & 58.89  & 54.67  & \cellcolor[rgb]{ .906,  .902,  .902}50.64  & \cellcolor[rgb]{ .886,  .937,  .855}\textbf{42.77 } & \cellcolor[rgb]{ .663,  .816,  .557}\textbf{32.87 } \\
          & RMSE  & 87.66  & 88.11  & 96.72  & 97.20  & 88.68  & 79.51  & 79.79  & 85.29  & 75.72  & 78.74  & 81.25  & 76.83  & \cellcolor[rgb]{ .906,  .902,  .902}76.32  & \cellcolor[rgb]{ .886,  .937,  .855}\textbf{60.64 } & \cellcolor[rgb]{ .663,  .816,  .557}\textbf{48.63 } \\
          & MAPE  & 39.97\% & 37.98\% & 47.71\% & 41.54\% & 35.00\% & 34.19\% & 30.75\% & 34.89\% & 29.13\% & 33.99\% & 35.89\% & 32.10\% & \cellcolor[rgb]{ .906,  .902,  .902}29.55\% & \cellcolor[rgb]{ .886,  .937,  .855}\textbf{23.47\%} & \cellcolor[rgb]{ .663,  .816,  .557}\textbf{19.99\%} \\
    \midrule
    \multirow{3}[2]{*}{PeMS08} & MAE   & 68.96  & 69.58  & 71.86  & 73.86  & 66.14  & 67.53  & 61.59  & 64.91  & 55.46  & 66.52  & 67.51  & 65.49  & \cellcolor[rgb]{ .906,  .902,  .902}58.77  & \cellcolor[rgb]{ .886,  .937,  .855}\textbf{30.08 } & \cellcolor[rgb]{ .663,  .816,  .557}\textbf{23.38 } \\
          & RMSE  & 95.19  & 95.19  & 96.32  & 106.21  & 91.67  & 89.72  & 86.26  & 88.83  & 81.35  & 89.08  & 91.35  & 87.47  & \cellcolor[rgb]{ .906,  .902,  .902}82.98  & \cellcolor[rgb]{ .886,  .937,  .855}\textbf{44.31 } & \cellcolor[rgb]{ .663,  .816,  .557}\textbf{35.46 } \\
          & MAPE  & 68.96\% & 68.66\% & 78.77\% & 76.36\% & 66.01\% & 65.58\% & 60.77\% & 63.96\% & 54.35\% & 64.41\% & 66.84\% & 63.21\% & \cellcolor[rgb]{ .906,  .902,  .902}58.24\% & \cellcolor[rgb]{ .886,  .937,  .855}\textbf{36.11\%} & \cellcolor[rgb]{ .663,  .816,  .557}\textbf{31.50\%} \\
    \bottomrule
    \end{tabular}%
    }
  \label{tab:DBFP}%
\end{table*}%

\subsection{Study on Fine Granularity}
\begin{figure}[t]
\captionsetup[subfloat]{font=footnotesize,labelfont=rm,textfont=rm}
\centering
\hspace{0.64cm}
\includegraphics[width=2.1in]{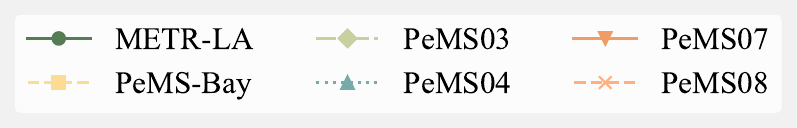} 
\newline
\subfloat[Performance of \( \text{STRP}_{avg} \) ]{\includegraphics[width=2.6in]{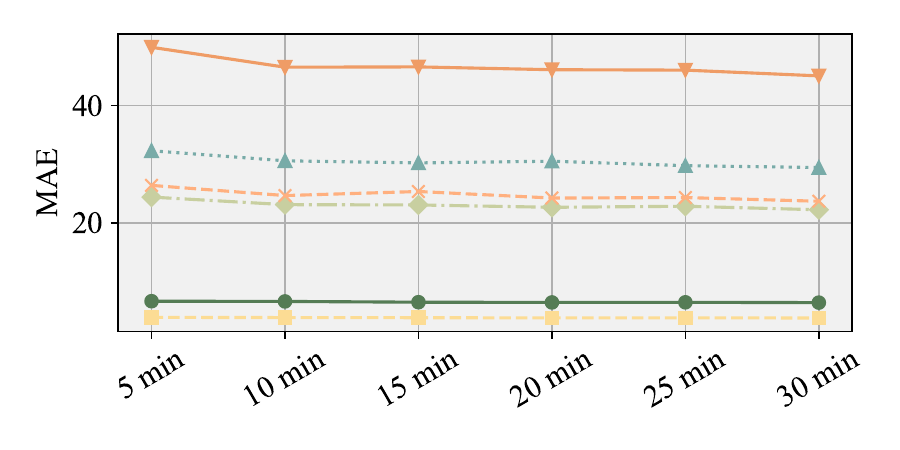}}
\newline
\subfloat[Performance of \( \text{STRP}_{att} \)]{\includegraphics[width=2.6in]{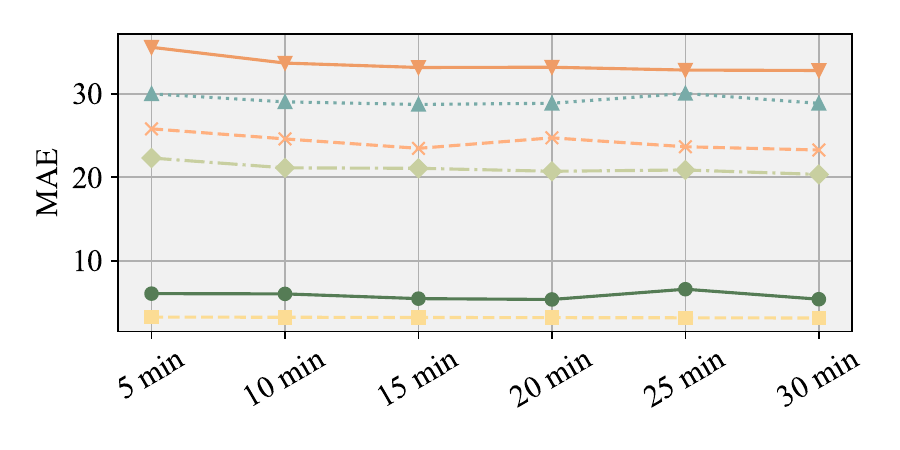}}
\newline
\caption{\label{fig:fine} Results of predicting the future at different fine granularities.}
\end{figure}

In previous experiments, we analyzed the fine-grained prediction capability of the STRP model for future $h$ = 5 minutes under different coarse-grained inputs. The aim of this fine-grained study is to investigate the performance of STRP in predicting future $D_p$ = 6 hours at granularities of $h$ = 5, 10, 15, 20, 25, and 30 minutes, given the coarse-grained conditions of $H$ = 30 minutes and $D_h$ = 6 hours. \textbf{Figure~\ref{fig:fine}} illustrates the performance of STRP$_{avg}$ (\textbf{subfigure (a)}) and STRP$_{att}$ (\textbf{subfigure (b)}). The results indicate that the STRP model shows improved prediction accuracy as h increases when predicting different granularities, which can be partly attributed to the fact that larger prediction granularities require fewer prediction steps within the same length of the prediction range. However, overall, the prediction results tend to stabilize, particularly on the METR-LA and PeMS-Bay datasets, where the reduction in error is not significant.

\subsection{Study on Coarse Granularity}
\begin{figure*}[ht]
 \centering
 \subfloat[The WBFP task for STRP\(_{avg} \) on METR-LA.]{\includegraphics[width=0.25\textwidth]{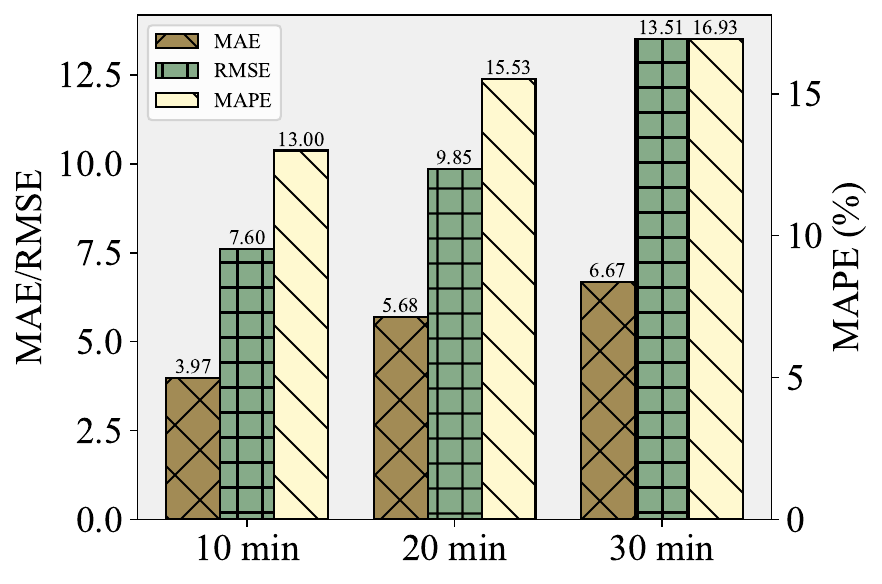}}\hfill
 \subfloat[The DBFP task for STRP\(_{avg} \) on METR-LA.]{\includegraphics[width=0.25\textwidth]{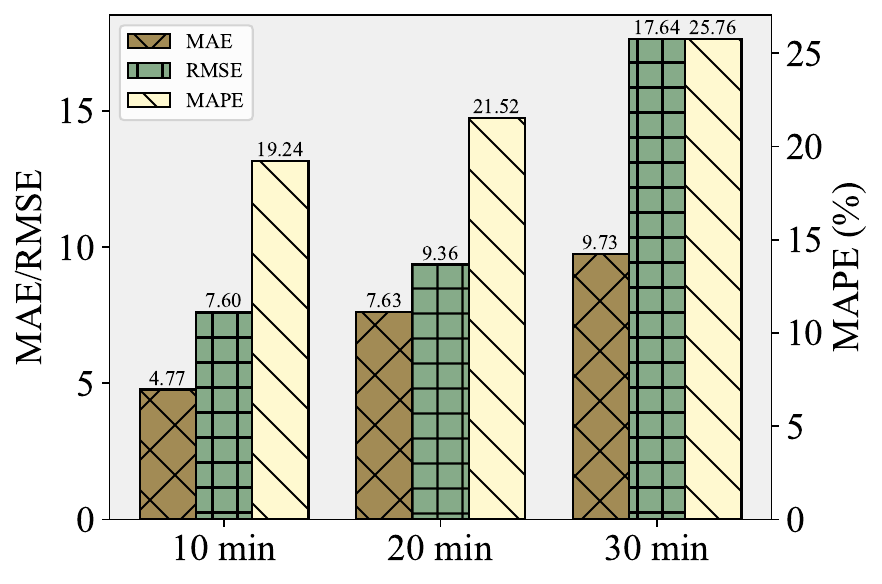}}\hfill
 \subfloat[The WBFP task for STRP\(_{att} \) on METR-LA.]{\includegraphics[width=0.25\textwidth]{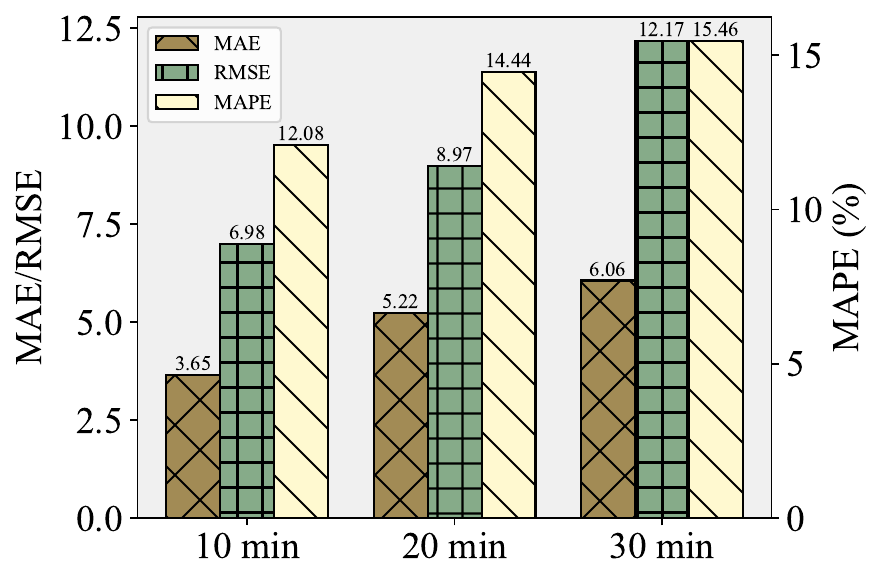}}\hfill
 \subfloat[The DBFP task for STRP\(_{att} \) on METR-LA.]{\includegraphics[width=0.25\textwidth]{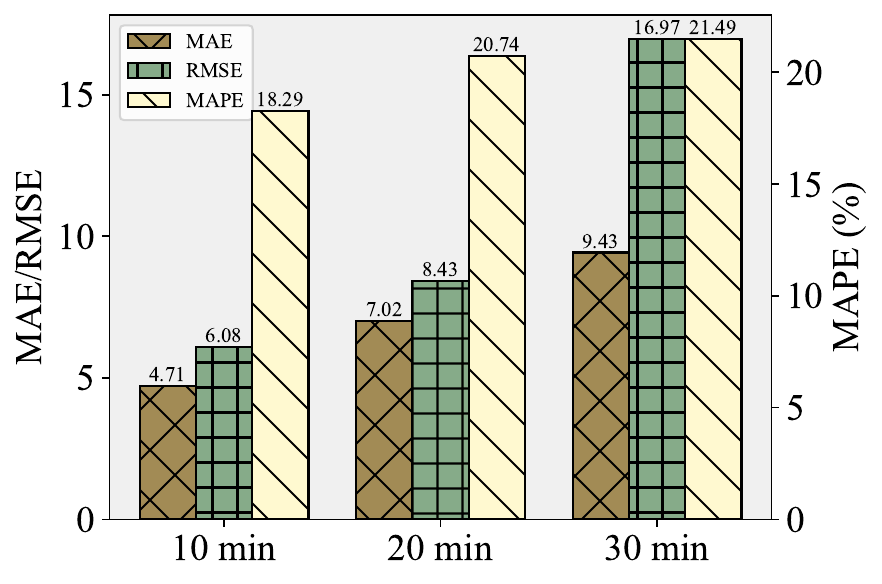}}
 \vspace{-0.3cm} %
 \subfloat[The WBFP task for STRP\(_{avg} \) on PeMS-Bay.]{\includegraphics[width=0.25\textwidth]{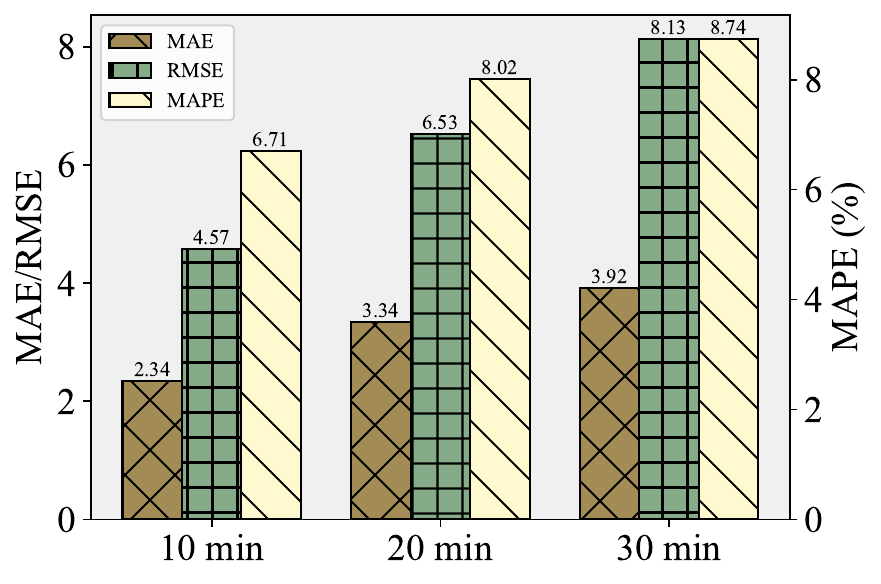}}\hfill
 \subfloat[The DBFP task for STRP\(_{avg} \) on PeMS-Bay.]{\includegraphics[width=0.25\textwidth]{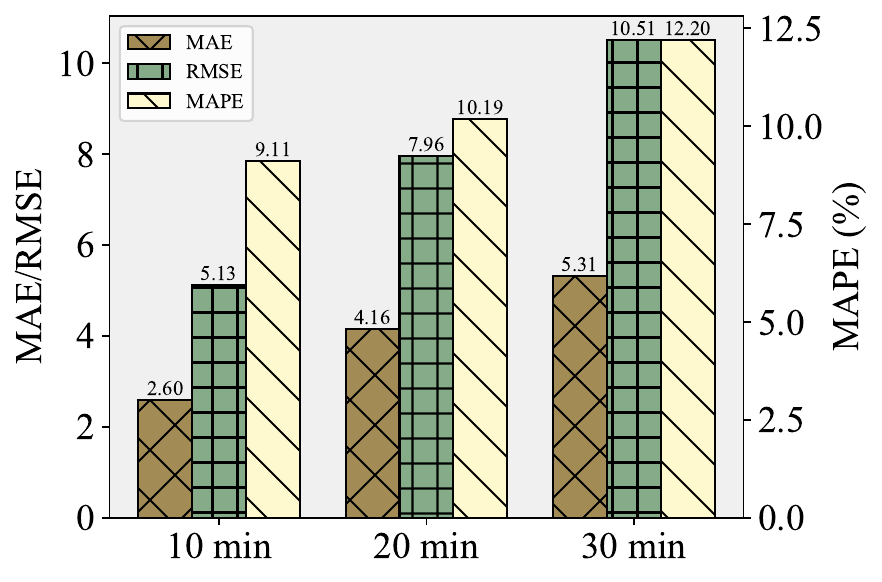}}\hfill
 \subfloat[The WBFP task for STRP\(_{att} \) on PeMS-Bay.]{\includegraphics[width=0.25\textwidth]{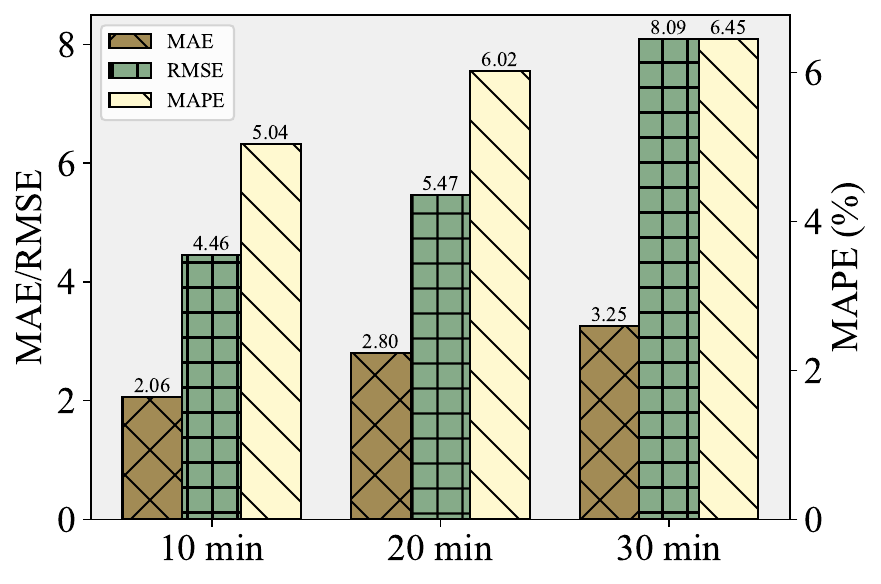}}\hfill
 \subfloat[The DBFP task for STRP\(_{att} \) on PeMS-Bay.]{\includegraphics[width=0.25\textwidth]{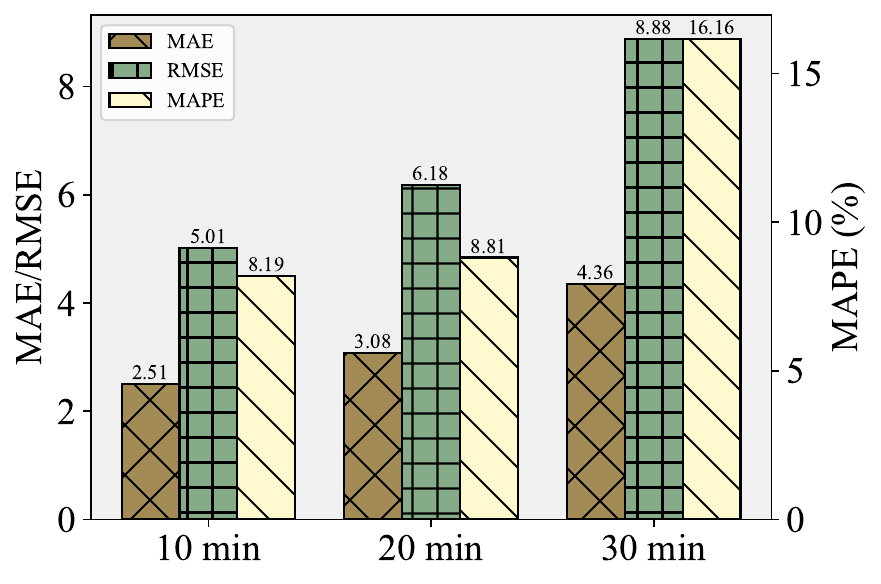}}
\vspace{-0.3cm} %
 \subfloat[The WBFP task for STRP\(_{avg} \) on PeMS03.]{\includegraphics[width=0.25\textwidth]{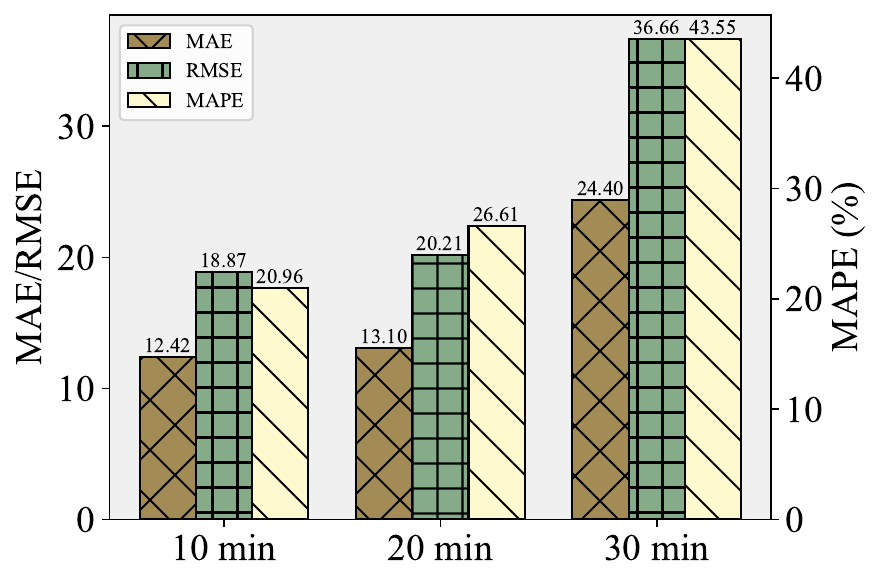}}\hfill
 \subfloat[The DBFP task for STRP\(_{avg} \) on PeMS03.]{\includegraphics[width=0.25\textwidth]{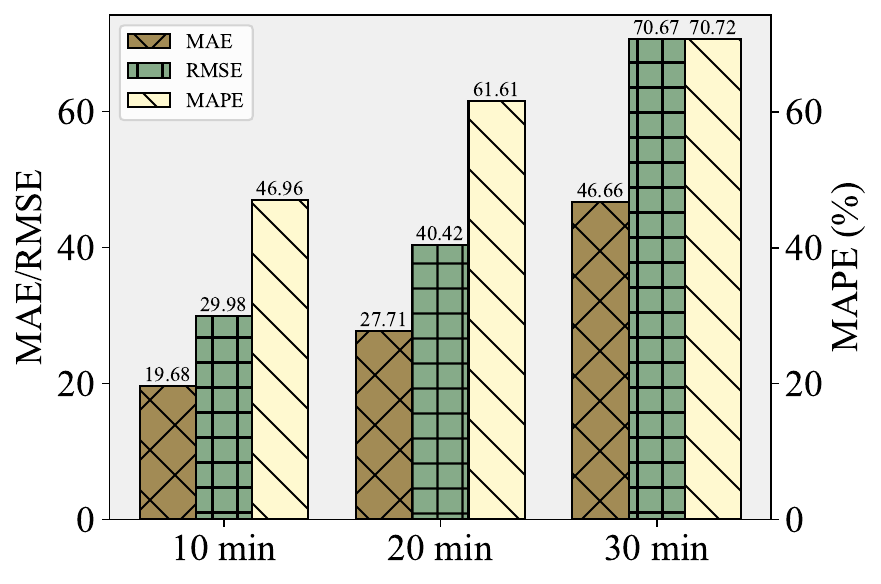}}\hfill
 \subfloat[The WBFP task for STRP\(_{att} \) on PeMS03.]{\includegraphics[width=0.25\textwidth]{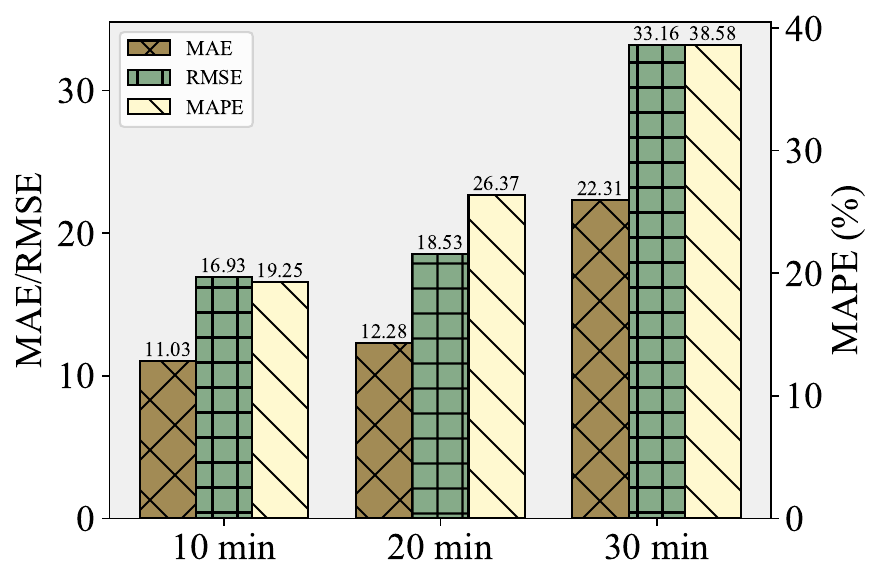}}\hfill
 \subfloat[The DBFP task for STRP\(_{att} \) on PeMS03.]{\includegraphics[width=0.25\textwidth]
 {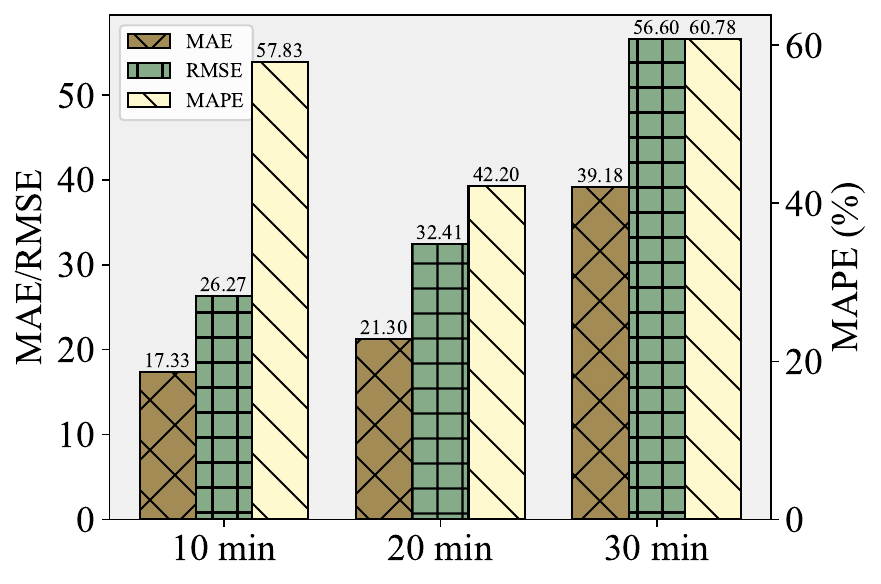}}
 \caption{\label{fig:coarse}Results of fine-grained future predictions $h= 5min$ using different coarse-grained historical data on the METR-LA, PeMS-Bay and PeMS03 datasets.}
\end{figure*}

To explore the ability to predict fine-grained future states under different coarse-grained data, we set three initial granularities: $H$ = 10 minutes, 20 minutes, and 30 minutes, to evaluate the performance of the two variants of the STRP model in WBFP and DBFP tasks. \textbf{Figure~\ref{fig:coarse}} presents the results across different granularities and tasks on the METR-LA, PeMS-Bay, and PeMS03 datasets; results on additional datasets are provided in the \textbf{supplementary material}. The findings clearly show that as the initial granularity becomes coarser, both STRP$_{avg}$ and STRP$_{att}$ exhibit higher prediction errors.  This suggests that finer-grained inputs provide richer information, while coarser inputs pose greater challenges as STRP must infer additional details. In extreme cases—for example, predicting from a 30-minute granularity to 5 minutes with a 1:6 generation ratio—the error amplifies to more than twice that of the 10-minute setting, underscoring the inherent difficulty of cross-granularity forecasting under extreme conditions. Moreover, STRP${att}$ consistently outperforms STRP${avg}$ in accuracy. WBFP also yields significantly lower errors than DBFP, reflecting the added difficulty of long-term forecasting. For DBFP, with fixed durations, larger granularity gaps require predicting more time steps, further highlighting the distinct challenges between the two tasks.

\begin{table}[htbp]
  \centering
  \caption{Ablation Study}
  \resizebox{\linewidth}{!}{
    \begin{tabular}{c|lll|lll}
    \toprule
    Variants & \multicolumn{3}{c|}{STRP$_{avg}$ w/o IDCov} & \multicolumn{3}{c}{STRP$_{att}$ w/o IDCov} \\
    \midrule
    Metrics & MAE & RMSE & MAPE & MAE & RMSE & MAPE \\
    \midrule
    METR-LA & 12.20(\textcolor{red!30}{\textuparrow 4.58}) & 14.97(\textcolor{red!30}{\textuparrow 5.61}) & 34.43\% (\textcolor{red!30}{\textuparrow 12.91\%}) & 10.89(\textcolor{red!30}{\textuparrow 3.86}) & 13.07(\textcolor{red!30}{\textuparrow 4.64}) & 32.15\% (\textcolor{red!30}{\textuparrow 11.41\%}) \\
    PeMS-Bay & 6.55(\textcolor{red!30}{\textuparrow 2.39}) & 12.44(\textcolor{red!30}{\textuparrow 4.48}) & 13.31\% (\textcolor{red!30}{\textuparrow 3.12\%}) & 4.81(\textcolor{red!30}{\textuparrow 1.73}) & 9.64(\textcolor{red!30}{\textuparrow 3.46}) & 13.75\% (\textcolor{red!30}{\textuparrow 4.93\%}) \\
    PeMS03 & 44.34(\textcolor{red!30}{\textuparrow 16.63}) & 64.67(\textcolor{red!30}{\textuparrow 24.25}) & 98.58\% (\textcolor{red!30}{\textuparrow 36.97\%}) & 32.58(\textcolor{red!30}{\textuparrow 11.29}) & 49.59(\textcolor{red!30}{\textuparrow 17.18}) & 64.57\% (\textcolor{red!30}{\textuparrow 22.37\%}) \\
    PeMS04 & 56.77(\textcolor{red!30}{\textuparrow 21.29}) & 81.71(\textcolor{red!30}{\textuparrow 30.64}) & 88.22\% (\textcolor{red!30}{\textuparrow 33.08\%}) & 42.77(\textcolor{red!30}{\textuparrow 14.82}) & 62.23(\textcolor{red!30}{\textuparrow 21.56}) & 67.78\% (\textcolor{red!30}{\textuparrow 23.48\%}) \\
    PeMS07 & 68.43(\textcolor{red!30}{\textuparrow 25.66}) & 97.02(\textcolor{red!30}{\textuparrow 36.38}) & 37.55\% (\textcolor{red!30}{\textuparrow 14.08\%}) & 50.94(\textcolor{red!30}{\textuparrow 18.08}) & 75.37(\textcolor{red!30}{\textuparrow 26.75}) & 30.98\% (\textcolor{red!30}{\textuparrow 10.99\%}) \\
    PeMS08 & 48.13(\textcolor{red!30}{\textuparrow 18.05}) & 70.89(\textcolor{red!30}{\textuparrow 26.58}) & 57.77\% (\textcolor{red!30}{\textuparrow 21.66\%}) & 36.00(\textcolor{red!30}{\textuparrow 12.62}) & 54.60(\textcolor{red!30}{\textuparrow 19.15}) & 48.52\% (\textcolor{red!30}{\textuparrow 17.01\%}) \\
    \midrule
    Variants & \multicolumn{3}{c|}{STRP w/o TreeCov} & \multicolumn{3}{c}{STGCN w TreeCov} \\
    \midrule
    Metrics & MAE & RMSE & MAPE & MAE & RMSE & MAPE \\
    \midrule
    METR-LA & 9.74(\textcolor{red!30}{\textuparrow 2.72}) & 11.83(\textcolor{red!30}{\textuparrow 3.40}) & 28.11\% (\textcolor{red!30}{\textuparrow 7.36\%}) & 10.99(\textcolor{green!50!black}{\textdownarrow 0.58}) & 13.35(\textcolor{green!50!black}{\textdownarrow 0.70}) & 29.80\% (\textcolor{green!50!black}{\textdownarrow 1.57\%}) \\
    PeMS-Bay & 4.81(\textcolor{red!30}{\textuparrow 1.73}) & 9.40(\textcolor{red!30}{\textuparrow 3.22}) & 12.64\% (\textcolor{red!30}{\textuparrow 3.83\%}) & 5.81(\textcolor{green!50!black}{\textdownarrow 0.31}) & 10.03(\textcolor{green!50!black}{\textdownarrow 0.53}) & 15.84\% (\textcolor{green!50!black}{\textdownarrow 0.83\%}) \\
    PeMS03 & 32.59(\textcolor{red!30}{\textuparrow 11.30}) & 48.43(\textcolor{red!30}{\textuparrow 16.02}) & 69.04\% (\textcolor{red!30}{\textuparrow 26.84\%}) & 89.87(\textcolor{green!50!black}{\textdownarrow 4.73}) & 125.89(\textcolor{green!50!black}{\textdownarrow 6.63}) & 175.28\% (\textcolor{green!50!black}{\textdownarrow 9.23\%}) \\
    PeMS04 & 42.19(\textcolor{red!30}{\textuparrow 14.23}) & 61.00(\textcolor{red!30}{\textuparrow 20.33}) & 66.13\% (\textcolor{red!30}{\textuparrow 21.83\%}) & 71.03(\textcolor{green!50!black}{\textdownarrow 3.74}) & 99.11(\textcolor{green!50!black}{\textdownarrow 5.22}) & 87.07\% (\textcolor{green!50!black}{\textdownarrow 4.58\%}) \\
    PeMS07 & 50.30(\textcolor{red!30}{\textuparrow 17.43}) & 72.66(\textcolor{red!30}{\textuparrow 24.03}) & 28.90\% (\textcolor{red!30}{\textuparrow 8.91\%}) & 59.36(\textcolor{green!50!black}{\textdownarrow 3.12}) & 83.71(\textcolor{green!50!black}{\textdownarrow 4.41}) & 36.08\% (\textcolor{green!50!black}{\textdownarrow 1.90\%}) \\
    PeMS08 & 35.55(\textcolor{red!30}{\textuparrow 12.17}) & 53.04(\textcolor{red!30}{\textuparrow 17.59}) & 44.96\% (\textcolor{red!30}{\textuparrow 13.46\%}) & 66.10(\textcolor{green!50!black}{\textdownarrow 3.48}) & 90.43(\textcolor{green!50!black}{\textdownarrow 4.76}) & 65.22\% (\textcolor{green!50!black}{\textdownarrow 3.43\%}) \\
    \bottomrule
    \end{tabular}%
  }
  \label{tab:ablation}%
\end{table}%

\subsection{Ablation Study}

To validate the contributions of the Tree Convolution module (TreeConv) and the Inverse Dilated Convolution module (IDConv) to the model, as well as the role of TreeConv as a plug-in that enhances other models, we designed a series of ablation experiments. These experiments included replacing IDConv with a linear layer under different pooling methods to complete fine-grained predictions; removing the Tree Convolution module and only using Inverse Dilated Convolution; and replacing the graph convolution in the STGCN model with TreeConv using attention pooling. \textbf{Table~\ref{tab:ablation}} uses different colors to indicate the corresponding changes in error (\textcolor{red!30}{increase \textuparrow} and \textcolor{green!50!black}{ decrease \textdownarrow}). The results show that removing the Tree Convolution module (TreeConv) and the Inverse Dilated Convolution module (IDConv) leads to varying degrees of increased prediction errors, particularly the removal of IDConv has a significant impact, causing errors to rise by 20\% to 30\%. This indicates that the unique design of Inverse Dilated Convolution is crucial for fine-grained prediction tasks. The design of TreeConv primarily aims to reduce the number of parameters, thereby expanding the deployment scenarios of the model. Nevertheless, the combination of STGCN and TreeConv still provides some performance improvement for STGCN.

\begin{figure}[t]
\captionsetup[subfloat]{labelfont=scriptsize,textfont=scriptsize}
\centering
\includegraphics[width=2.6in]{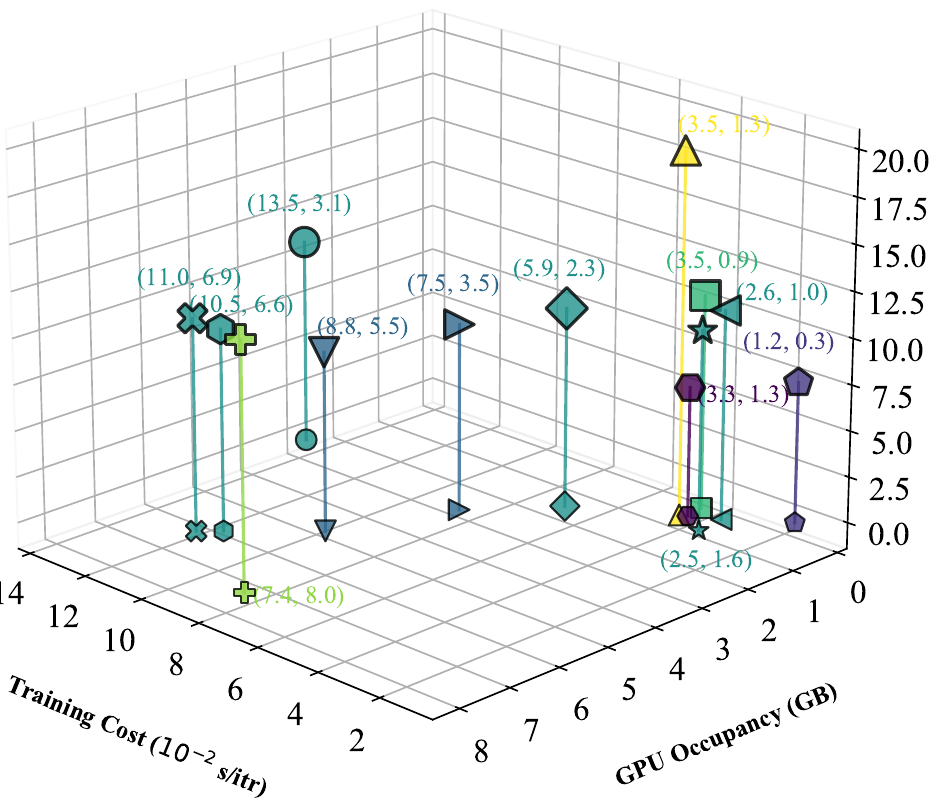}
\includegraphics[width=0.7in]{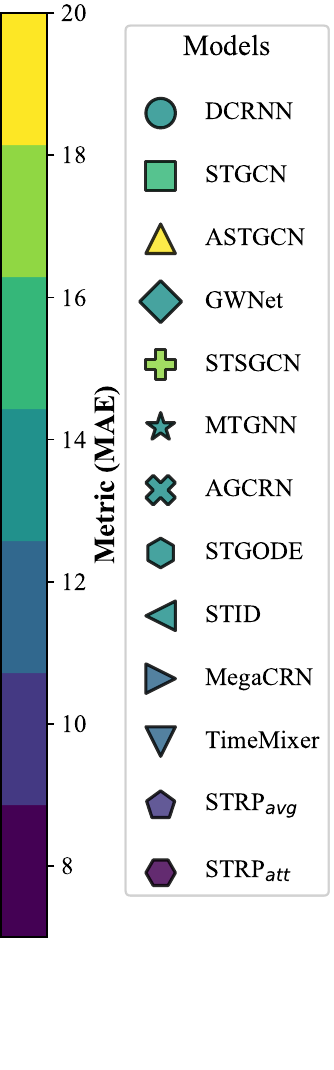} 
\caption{\label{fig:cost} Comparison of training cost and resource requirements (Training Cost, GPU Occupancy), with color and line length representing the magnitude of MAE.} 
\end{figure}
\subsection{Cost and Performance Analysis}
To evaluate the cost-effectiveness of STRP in resource-constrained scenarios, we tested its performance on the METR-LA dataset with $H=20\text{min}, h=20\text{min}$. The experiments were conducted on a device equipped with an Intel(R) Xeon(R) Gold 6278C CPU @ 2.60GHz and an NVIDIA GeForce RTX 2080 GPU. We compared STRP with baseline models in terms of training time per iteration, GPU memory usage, and MAE (note: STAEformer and Trafformer could not be deployed due to excessive memory requirements). As shown in \textbf{Figure~\ref{fig:cost}}, STRP${avg}$ demonstrated excellent efficiency, requiring only $1.2 \times 10^{-2}$ seconds per iteration and 0.3GB GPU memory while maintaining competitive accuracy. By contrast, STRP${att}$ achieved the best accuracy but with slightly higher time and memory costs than the simpler STID model. Among baselines, MTGNN, STGCN, and ASTGCN also showed deployment advantages due to lower resource demands.
Overall, STRP${avg}$ is the most cost-effective option in resource-limited settings, while STRP${att}$ is preferred when accuracy is prioritized. In practice, we recommend using STRP${avg}$ for edge-device or real-time inference, and STRP${att}$ when computational resources are sufficient and high accuracy is required, enabling a balanced trade-off between efficiency and performance.

\subsection{Interpretability Analysis}

\begin{figure}[h]
\vspace{-0.25cm}
\captionsetup[subfloat]{font=footnotesize,labelfont=rm,textfont=rm}
\centering
\subfloat[STRP$_{att}$ heatmap]{
  \includegraphics[width=1.7in]{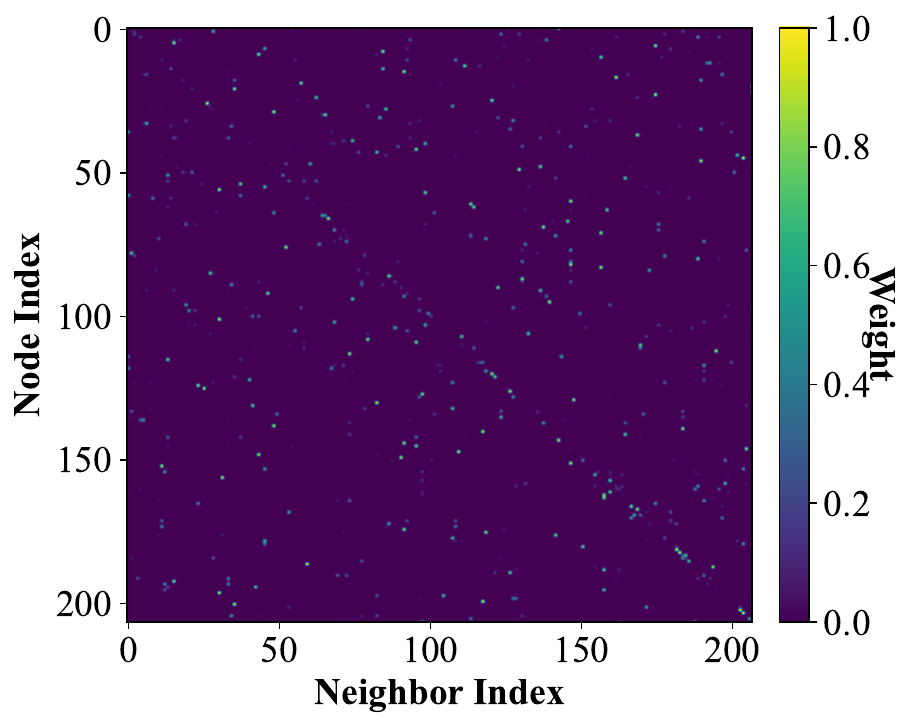}
}
\subfloat[STRP$_{avg}$ heatmap]{
  \includegraphics[width=1.7in]{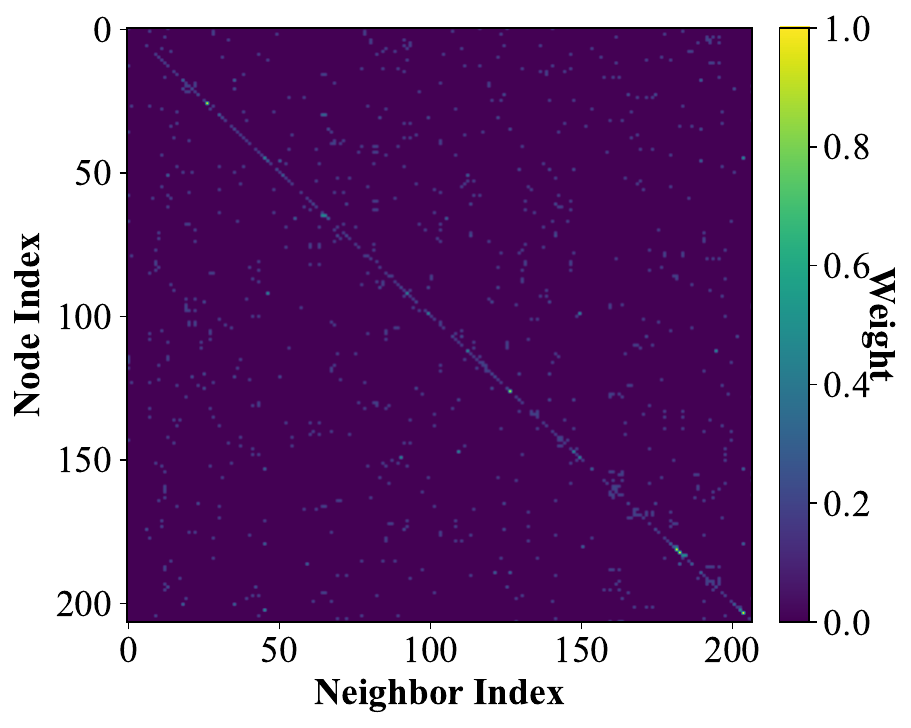}
}
\caption{\label{fig:heatmap_metrla} Visualization of node-contribution heatmaps for STRP$_{att}$ and STRP$_{avg}$.}
\end{figure}
To assess the interpretability of STRP, we visualize the two aggregation strategies in the TreeConv module on the METR-LA dataset, with node-contribution heatmaps shown in \textbf{Figure~\ref{fig:heatmap_metrla}}. For STRP$_{att}$, the heatmap reveals clear heterogeneity in neighbor contributions, where certain key nodes consistently receive higher weights across time, indicating that the attention mechanism can highlight spatial dependencies most relevant for prediction. In contrast, STRP$_{avg}$ produces a more uniform distribution, reflecting its equal-weight pooling design that emphasizes efficiency and robustness. The comparison demonstrates their complementary roles: STRP$_{att}$ captures fine-grained critical dependencies, while STRP$_{avg}$ provides stable and cost-effective aggregation, offering practical guidance for model selection under different application needs.

\section{Related Works}
\subsection{Graph Convolutional Networks (GCNs) \label{related:A}}
GCNs are designed to process non-Euclidean data, achieving significant success in fields like traffic prediction and social network analysis. Initially, graph convolution was defined in the spectral domain \cite{bruna2014spectral} using the eigen-decomposition of the graph Laplacian:
\begin{equation}
    \mathbf{y} = \mathbf{U} \mathbf{g}_{\theta} \mathbf{U}^T \mathbf{x},
\end{equation}
where \(\mathbf{U}\) is the eigenvector matrix of the Laplacian, \(\mathbf{g}_{\theta}\) denotes the filter parameters, and \(\mathbf{x}\) represents the input signal. While mathematically rigorous, this approach is computationally impractical for large-scale graphs.

To enhance efficiency, ChebNet introduced Chebyshev polynomials to approximate graph convolution filters \cite{defferrard2016convolutional}, avoiding the costly eigen-decomposition. Later, GCN further simplified this by adopting a first-order approximation \cite{kipf2016semi}:

\begin{equation}
    \mathbf{H}^{(l+1)} = \sigma \left( \mathbf{\tilde{D}}^{-\frac{1}{2}} \mathbf{\tilde{A}} \mathbf{\tilde{D}}^{-\frac{1}{2}} \mathbf{H}^{(l)} \mathbf{W}^{(l)} \right),
\end{equation}
where \(\mathbf{\tilde{A}}\) is the adjacency matrix with self-loops, \(\mathbf{W}^{(l)}\) is the weight matrix, and \(\sigma\) is an activation function. This simplification reduced computational complexity significantly.

Graph Attention Networks (GATs) introduced attention mechanisms \cite{velivckovic2017graph} to dynamically adjust aggregation weights between nodes:

\begin{equation}
    \alpha_{ij} = \frac{\exp(e_{ij})}{\sum_{k \in \mathcal{N}(i)} \exp(e_{ik})}, \quad \mathbf{h}_i^{(l+1)} = \sigma \left( \sum_{j \in \mathcal{N}(i)} \alpha_{ij} \mathbf{W} \mathbf{h}_j^{(l)} \right),
\end{equation}
where \(\alpha_{ij}\) represents attention weights, allowing for adaptive importance assignment, thereby enhancing the model's representational power. Despite the significant advancements of GCNs, they still suffer from high computational overhead and the well-known issue of feature over-smoothing \cite{pope2019explainability,song2020stronger}. To address scalability challenges on large graphs and high-dimensional data, a wide range of methods have been proposed to simplify computation, including first-order approximations \cite{kipf2016semi}, neighbor sampling \cite{hamilton2017inductive,zeng2019graphsaint}, sparse message passing \cite{wu2019simplifying, rong2019dropedge}, and batch-parallel optimization \cite{chiang2019cluster}. However, due to their reliance on full-graph adjacency matrices for aggregation, GCNs often produce non-unique message-passing paths, making it difficult to trace how node representations are formed, thereby limiting their interpretability. In contrast, trees represent a structurally simplified subset of graphs, and their potential for interpretable and efficient modeling remains underexplored.

\subsection{Traffic Prediction}

Traffic prediction has been a key research focus due to its relevance to daily life and advancements in the Internet of Vehicles\cite{kumar2015short,avila2020data}. Accurate prediction hinges on spatio-temporal modeling, applied across different scales of traffic scenarios, including city-level \cite{jin2022selective,mo2022cross}, regional-level \cite{cui2023roi,li2023st}, and lane-level contexts \cite{li2024st}.

Early methods used Convolutional Neural Networks (CNNs) to model spatial features by converting traffic networks into image grids \cite{li2018brief}. However, CNNs are limited to Euclidean spaces and struggle with complex topologies. GCNs later became mainstream for non-Euclidean data modeling \cite{li2017diffusion,yu2017spatio,wu2019graph,fang2021spatial}, especially when enhanced by attention mechanisms to capture dynamic traffic features \cite{guo2019attention,geng2024stgaformer}. TreeCN\footnote{TreeCN was not included as a baseline due to code incompatibility.} introduced tree structures to model spatial directionality for the first time but retained GCN's matrix-based principles, failing to fully overcome its limitations\cite{lv2023treecn}. For temporal modeling, Recurrent Neural Networks (RNNs) replaced early fully convolutional models for capturing short-term dynamics \cite{bai2020adaptive,li2018csrnet,yu2015multi}. More recently, Transformer models \cite{jin2023trafformer,jiang2023pdformer,vaswani2017attention}, with their superior ability to model long-term dependencies, have gradually replaced RNNs. Additionally, dilated causal convolutions and MLP-based models have enhanced parallelism and computational efficiency\cite{wu2019graph,shao2022spatial}.

While multi-scale and multi-horizon temporal modeling has been emphasized \cite{fan2019multi,lim2021temporal}, existing methods are constrained by training-data granularity, restricting predictions to fixed resolutions and frequencies. Recent latent diffusion models—LDM4TS \cite{ruan2025vision}, MTSCI \cite{zhou2024mtsci}, and mr-Diff \cite{shen2024multi}—show strong generative and multi-scale capacity, while continuous-time approaches such as MNDE \cite{liu2024multi} and the NDE \cite{oh2025comprehensive} offer new insights for irregular sampling and dynamic modeling. However, these works mainly address generative or continuous dynamics, whereas STRP targets cross-granularity forecasting, complementing these directions.

\section{Complexity Analysis}

We provide a detailed analysis of the time and space complexity of STRP and compare it with traditional GCNs.

\subsection{Complexity of STRP}

\subsubsection{Time Complexity}

The STRP model consists of two core modules: the Tree Convolution Module and the Inverse Dilated Convolution Module. 

For the tree convolution module, the first step is constructing the adjacency list and building the hierarchical tree. Given a graph with \( N \) nodes and average degree \( d \), the construction cost is \( O(N \cdot d) \), which simplifies to \( O(N) \) as \( d \ll N \) in sparse graphs. Then, for the bottom-up hierarchical convolution, each node aggregates feature representations from its children. At each layer, a linear transformation of feature dimension \( F \) is applied, resulting in a per-layer cost of \( O(N \cdot F) \). Assuming the tree depth is \( L \approx \log N \), the total complexity of the tree convolution module becomes \( O(N \cdot F \cdot \log N) \).

For the inverse dilated convolution module, at each time step, the diffusion kernel computes across a fixed temporal window. Given sequence length \( T \), kernel size \( K \), and dilation stride \( d \), each step requires \( O(K \cdot d) \), which simplifies to \( O(1) \) under constant kernel settings. Applying this across all \( T \) time steps over \( L \) refinement layers leads to an overall complexity of \( O(T \cdot L) \).

Combining both modules, the total time complexity of STRP is:
\begin{equation}
O(N \cdot F \cdot \log N) + O(T \cdot L).
\end{equation}

\subsubsection{Space Complexity}

For space complexity, the tree convolution module requires \( O(N \cdot d) \approx O(N) \) to store the adjacency structure. It also requires \( O(L \cdot N \cdot F) \) to store the intermediate node features over \( L \) layers with feature dimension \( F \).

In the inverse dilated convolution module, storing the convolution kernel parameters across \( L \) layers with kernel size \( K \cdot d \) leads to \( O(L \cdot K \cdot d) \), which is small and treated as \( O(L) \) in practice. In addition, each layer must store feature maps of size \( T \cdot F \), leading to a total of \( O(L \cdot T \cdot F) \).

Hence, the overall space complexity of STRP is:
\begin{equation}
O(N) + O(L \cdot N \cdot F) + O(L \cdot T \cdot F).
\end{equation}

\subsection{Complexity of GCNs}

\subsubsection{Time Complexity}

In traditional GCNs, each node aggregates information from its neighbors based on the full graph adjacency. Given \( N \) nodes, \( E \) edges, and feature dimension \( F \rightarrow F' \), each layer requires \( O(E \cdot F) \) for adjacency-feature multiplication and \( O(N \cdot F \cdot F') \) for linear transformation. For a network with \( L \) layers, the total time complexity is:
\begin{equation}
O(L \cdot (E \cdot F + N \cdot F \cdot F')).
\end{equation}

\subsubsection{Space Complexity}

The adjacency matrix requires \( O(E) \) space when stored in a sparse format. Additionally, feature storage across \( L \) layers leads to \( O(L \cdot N \cdot F) \). Therefore, the total space complexity of GCNs is:
\begin{equation}
O(E) + O(L \cdot N \cdot F).
\end{equation}

\subsection{Comparison between TreeConv and GCNs}

In terms of time complexity, tree convolution achieves \( O(N \cdot F \cdot \log N) \) due to its hierarchical structure and localized feature aggregation, whereas GCN incurs \( O(L \cdot (E \cdot F + N \cdot F \cdot F')) \) in total, with each layer performing full-neighborhood aggregation and transformation. The cost of GCN becomes particularly high in dense graphs where \( E \gg N \), while TreeConv scales more favorably with graph size and feature dimension. From a space perspective, tree convolution requires only \( O(N) \) space for storing simplified hierarchical adjacency, while GCN needs \( O(E) \) space for full edge storage. Both approaches require \( O(L \cdot N \cdot F) \) for storing multi-layer node features. Overall, STRP provides significant computational and memory advantages, especially in large-scale spatio-temporal prediction tasks.

\subsection{Parameter Efficiency and Accuracy}

\begin{figure}[t]
\centerline{\includegraphics[width=0.5\textwidth]{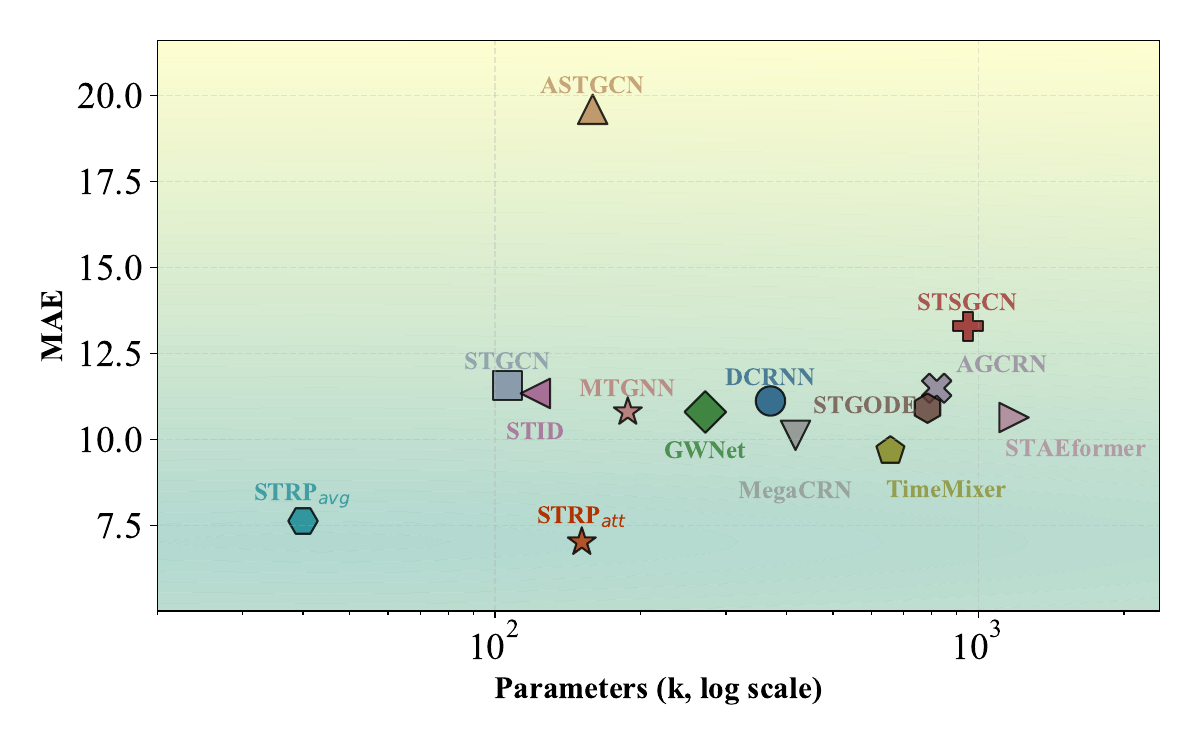}}
\caption{Comparison of model size vs. MAE on METR-LA (DBFP, 20min); lower left indicates better trade-off.}
\label{fig:param}
\end{figure}

We further evaluate the efficiency of STRP by comparing prediction accuracy and parameter size in Figure~\ref{fig:param}. STRP$_\text{att}$ achieves the best accuracy (MAE = 7.02) with 151k parameters. STRP$_\text{avg}$ maintains competitive accuracy with only 40k parameters, outperforming larger baselines like STGCN and ASTGCN. These results confirm that STRP offers superior cost-effectiveness and is well-suited for deployment in resource-constrained environments.
\section{Conclusions and Future Works}
This paper introduces a novel problem in spatio-temporal data management: inferring fine-grained future traffic states from coarse-grained historical data. We formalize this problem into two sub-tasks—Window-Based Fine-Grained Prediction (WBFP) and Duration-Based Fine-Grained Prediction (DBFP)—with the goal of enriching the utility of coarse-grained historical traffic data while reducing storage overhead in database systems. To address this challenge, we propose STRP, a lightweight yet accurate prediction framework that integrates Tree Convolution and Inverse Dilated Convolution. Tree Convolution hierarchically aggregates spatial information through a tree structure, employing average pooling for efficiency and attention pooling for expressive modeling. Inverse Dilated Convolution incrementally expands the temporal resolution to recover fine-grained dynamics from coarse inputs. Extensive experiments demonstrate the effectiveness of STRP in fine-grained prediction tasks and highlight the unique challenges posed by temporal granularity mismatch. Future work will focus on improving the scalability of Tree Convolution and extending fine-grained prediction capabilities to broader spatio-temporal data management scenarios.



\bibliographystyle{IEEEtran}
\bibliography{ref}

\end{document}